\documentclass{article}

\usepackage[preprint]{neurips_2025}

\usepackage[utf8]{inputenc} 
\usepackage[T1]{fontenc}    
\usepackage{hyperref}       
\usepackage{url}            
\usepackage{booktabs}       
\usepackage{amsfonts}       
\usepackage{nicefrac}       
\usepackage{microtype}      
\usepackage[dvipsnames]{xcolor}         

\usepackage{wrapfig}        
\usepackage{graphicx}
\usepackage{enumitem}
\usepackage{caption}

\usepackage{amsmath,bm}
\usepackage{dsfont}
\usepackage{amssymb}
\usepackage{amsfonts}
\usepackage{dcolumn} 
\usepackage{tabu}
\usepackage{array}
\usepackage{xspace}
\usepackage{makecell}

\usepackage{colortbl}
\usepackage{booktabs, multirow} 
\usepackage{soul}
\usepackage{changepage,threeparttable} 

\PassOptionsToPackage{numbers,compress,sort}{natbib}
\usepackage[numbers]{natbib}

\usepackage{algorithm}
\usepackage{algpseudocode}
\usepackage[vlined,linesnumbered,ruled,algo2e]{algorithm2e}

\usepackage{amsmath}
\usepackage{amssymb}
\usepackage{mathtools}
\usepackage{amsthm}
\usepackage{subcaption}
\usepackage[capitalize,noabbrev]{cleveref}

\theoremstyle{plain}

\theoremstyle{definition}

\theoremstyle{remark}

\newcommand{\simulator}{s}
\newcommand{\encoder}{\textsc{Encoder}}
\newcommand{\processor}{\textsc{Processor}}
\newcommand{\decoder}{\textsc{Decoder}}

\newcommand{\norm}[1]{\lVert#1\rVert}

\title{Hybrid Neural-MPM for Interactive Fluid Simulations\\in Real-Time}

\newcommand*\samethanks[1][\value{footnote}]{\footnotemark[#1]}
\author{
  Jingxuan Xu\thanks{Equal Contribution.}\\
  Beijing Jiaotong University\\
  \And
  Hong Huang\samethanks\\
  Simon Fraser University\\
  \And
  Chuhang Zou\\
  Meta Reality Labs\\
  \And
  Manolis Savva\\
  Simon Fraser University\\
  \And
  Yunchao Wei\thanks{Co-corresponding authors.}\\
  Beijing Jiaotong University\\
  \And
  Wuyang Chen\samethanks\\
  Simon Fraser University\\
}

\begin{document}

\maketitle

\begin{abstract}
We propose a neural physics system for real-time, interactive fluid simulations.
Traditional physics-based methods, while accurate, are computationally intensive and suffer from latency issues.
Recent machine‑learning methods reduce computational costs while preserving fidelity; yet most still fail to satisfy the latency constraints for real‑time use and lack support for interactive applications.
To bridge this gap, we introduce a novel hybrid method that integrates numerical simulation, neural physics, and generative control.
Our neural physics jointly pursues low-latency simulation and high physical fidelity by employing
a fallback safeguard to classical numerical solvers.
Furthermore, we develop a diffusion-based controller that is trained using a reverse modeling strategy to generate external dynamic force fields for fluid manipulation.
Our system demonstrates robust performance across diverse 2D/3D scenarios, material types, and obstacle interactions, achieving real-time simulations at high frame rates ({$-11\sim 29$\% latency}) while
enabling fluid control guided by
user-friendly freehand sketches.
We present a significant step towards practical, controllable, and physically plausible fluid simulations for real-time interactive applications.
We will release our code, model, and data at \href{https://hybridmpm.github.io/}{\emph{https://hybridmpm.github.io/}}.
\end{abstract}

\begin{figure}[h!]
\vspace{-0.5em}
	\centering
	\includegraphics[width=0.85\textwidth]{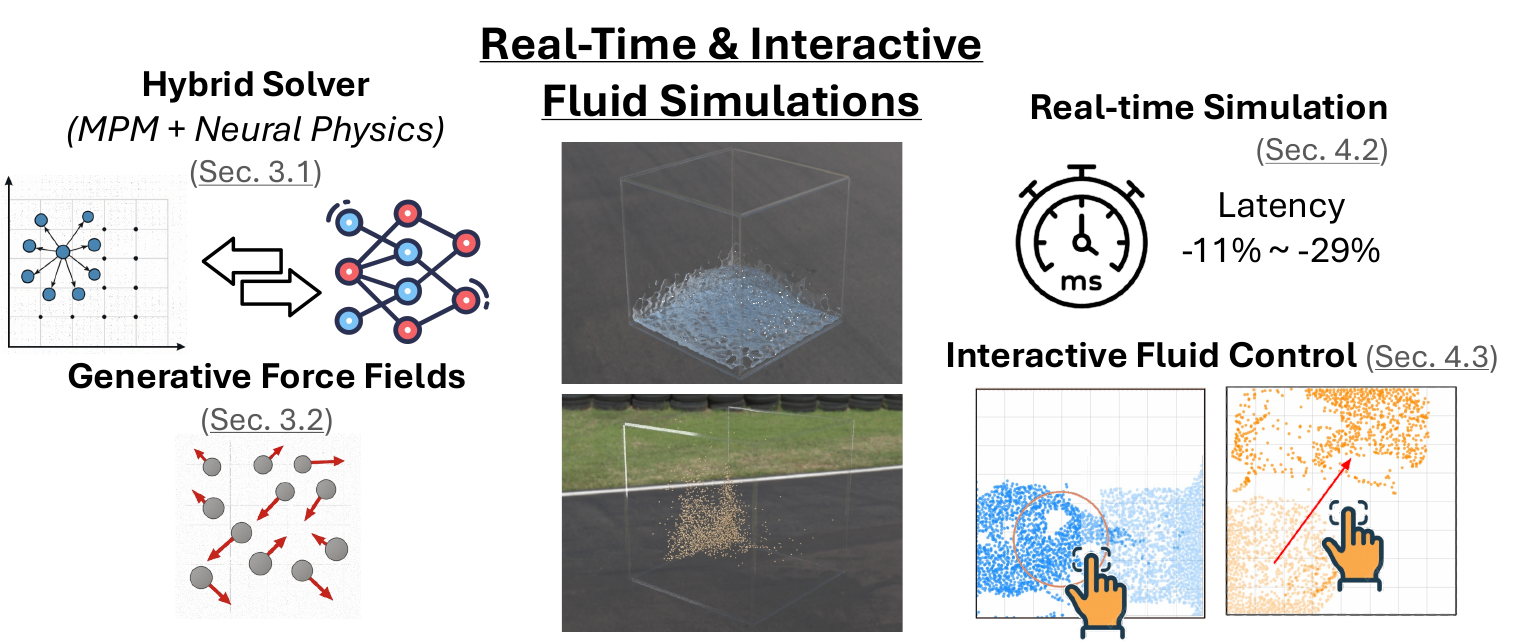}
    \vspace{-0.5em}
    \captionsetup{font=small}
    \caption{We target real-time, interactive fluid simulations.
Our hybrid solver integrates a numerical simulator and neural physics (Section \ref{sec:realtime_hybrid}), enabling real-time simulation (Section \ref{sec:exp_acceleration}).
In addition, we generate external force fields (Section \ref{sec:interactive_control}) to support users to control fluids interactively via freehand sketches (Section \ref{sec:exp_control}).
    }
    \label{fig:teaser}
    \vspace{-.5em}
\end{figure}

\section{Introduction}

Modeling fluid behavior is essential for advancing diverse engineering fields,
including
entertainment~\cite{stam2023stable}, urban planning~\cite{blocken2013cfd}, fashion design~\cite{volino2005early}, and virtual reality (VR)~\cite{solmaz2022interactive}.
Moreover, controllability, aiming to instruct movements and shapes of fluids, is also a very important attribute for volumetric effects, character animations, and fluid-solid coupling~\cite{raveendran2012controlling}.
Realizing compelling and interactive physics simulations in real-time has been the long-standing objective
for years in order to deliver transformative user experiences.

Traditional simulation methods, though powerful, often demand significant implementation efforts and computational costs~\cite{bridson2015fluid}.
Recent neural physics and machine learning approaches present a promising path forward by learning from data,
delivering transformative changes
for use cases such as fluid interactions and animations~\cite{sanchez2020learning}.
However, fidelity and latency in these neural-based methods are not well-balanced.
Moreover, most methods only focus on the accuracy of non-interactive applications, and their computational complexity still remains generally high for real-time scenarios~\cite{brandstetter2022message}.

Motivated by the above challenges, we ask two scientific questions:

\vspace{-0.5em}
\begin{center}
\fbox{
    \parbox{0.9\linewidth}{
        \textit{\textbf{Q1}: Can neural physics accelerate real-time fluid simulations and interactions?}
        \newline
        \textit{\textbf{Q2}: Can neural physics and generative methods be optimized for interactive fluid control?}
    }
}
\end{center}
\vspace{-0.5em}

We aim to explore a novel paradigm: neural physics for interactive simulations in real-time (Figure~\ref{fig:teaser}).
We provide affirmative answers.
The core idea is to \textbf{proactively marry the strengths of numerical simulation (high fidelity), neural physics (low latency), and generative control (interactivity)} to deliver authentic and diverse fluid simulations.
Specifically, neural physics is responsible for significantly low-latency fluid simulation with tolerant errors, and numerical simulation will serve as a fallback solution when fluid dynamics is increasingly complex.
Furthermore, to make fluid animation compatible with user-friendly control, we introduce another diffusion-based controller to generate external force fields to assist manipulations.
We summarize our contributions below:
\begin{enumerate}[leftmargin=*]
    \item
    We improve the \textbf{error-latency trade-off} of fluid simulation.
    First, to accelerate neural physics, we seek to build our graph neural network at low spatiotemporal resolution without substantial degradation in simulation accuracy (Section~\ref{sec:low_spatiotemporal}).
    Second, to preserve simulation fidelity and avoid error accumulation during unrolling, we make our neural physics hybrid with a safeguard condition and fallback mechanism to the classic MPM (Material Point Method) algorithm (Section~\ref{sec:hybrid_solver}).

    \item
    We further aim to support \textbf{users' flexible freehand sketches} that specify desired trajectories or shapes of fluid particles to be controlled. To this end, our novel reverse simulation strategy enables the automated generation of realistic fluid control data (Section~\ref{sec:control_data}), which is used to train our diffusion-based generative controller (Section~\ref{sec:controlnet}).

    \item
    Across \textbf{diverse scenarios} (2D/3D, materials, rigid obstacles, see Table~\ref{tab:datasets}), our hybrid simulator can \textbf{significantly accelerate simulations} ({$-11\sim 29$\% latency}) while maintaining low errors (Section~\ref{sec:exp_acceleration}), and can \textbf{control fluid particles to align with user sketches} (Section~\ref{sec:exp_control}),
    paving the way for promising advances towards engaging interactive simulations in real-time.
\end{enumerate}

\section{Background}
\label{sec:background_gnn}

We first introduce the necessary components on which our method is built, and {how they can be made real-time and controllable in Section ~\ref{sec:methods}.}

\subsection{Fluid Simulations with Material Point Method (MPM)}

The Material Point Method (MPM)~\cite{jiang2015affine,hu2019taichi,hu2019difftaichi,hu2021quantaichi} is a hybrid Eulerian-Lagrangian numerical technique for simulating complex interactions between solid and fluid materials, especially under large deformations and topological changes (snow, landslides, cloth, etc.).
It extends the FLuids-Implicit-Particle (FLIP)~\cite{brackbill1986flip} from Computational Fluid Dynamics (CFD) to solid mechanics by representing materials as a set of Lagrangian particles that carry mass, velocity $\left(\dot{\bm{p}}_{i,t}\right)$, position $\left(\bm{p}_{i,t}\right)$,
and possible internal states.
These particle quantities are first transferred to a background Eulerian grid using a particle-to-grid mapping ($\mathrm{p2g}$).
The equations of motion are then solved on this grid,
after which updated values are mapped back to particles through grid-to-particle transfer ($\mathrm{g2p}$).
The particle positions ($\bm{p}$) are then advanced using the updated velocities ($\dot{\bm{p}}$), e.g., $\bm{p}_{i,t+1}=\bm{p}_{i,t}+\Delta t \cdot \dot{\bm{p}}_{i,t+1}$.

\subsection{GNN-based Neural Physics for Particle Simulations}
\label{background:neural_physics}

We denote the state of particle $i$ at time step $t$ as $\bm{x}_{i,t}$  (position $\bm{p}$, velocity $\dot{\bm{p}}$, acceleration $\ddot{\bm{p}}$, etc.),
and the state of $N$ particles as $X_t = [\bm{x}_{1,t}, \dots, \bm{x}_{N,t}]$.
A \textit{simulator} $\simulator$
maps $T_\text{in}$ input states to causally consequent future states,
and can iteratively compute ${X_{t_{T_\text{in}+1}} = \simulator(X_{t_1}, X_{t_2}, \cdots, X_{t_{T_\text{in}}})}$
to simulate a rollout trajectory.
Following~\cite{sanchez2020learning}, our learnable simulator $\simulator_\theta$ adopts a particle-based representation of the physical system,
which can be viewed as message-passing via a graph neural network (GNN).

\begin{wrapfigure}{r}{68mm}
\vspace{-2em}
	\centering
	\includegraphics[width=0.5\textwidth]{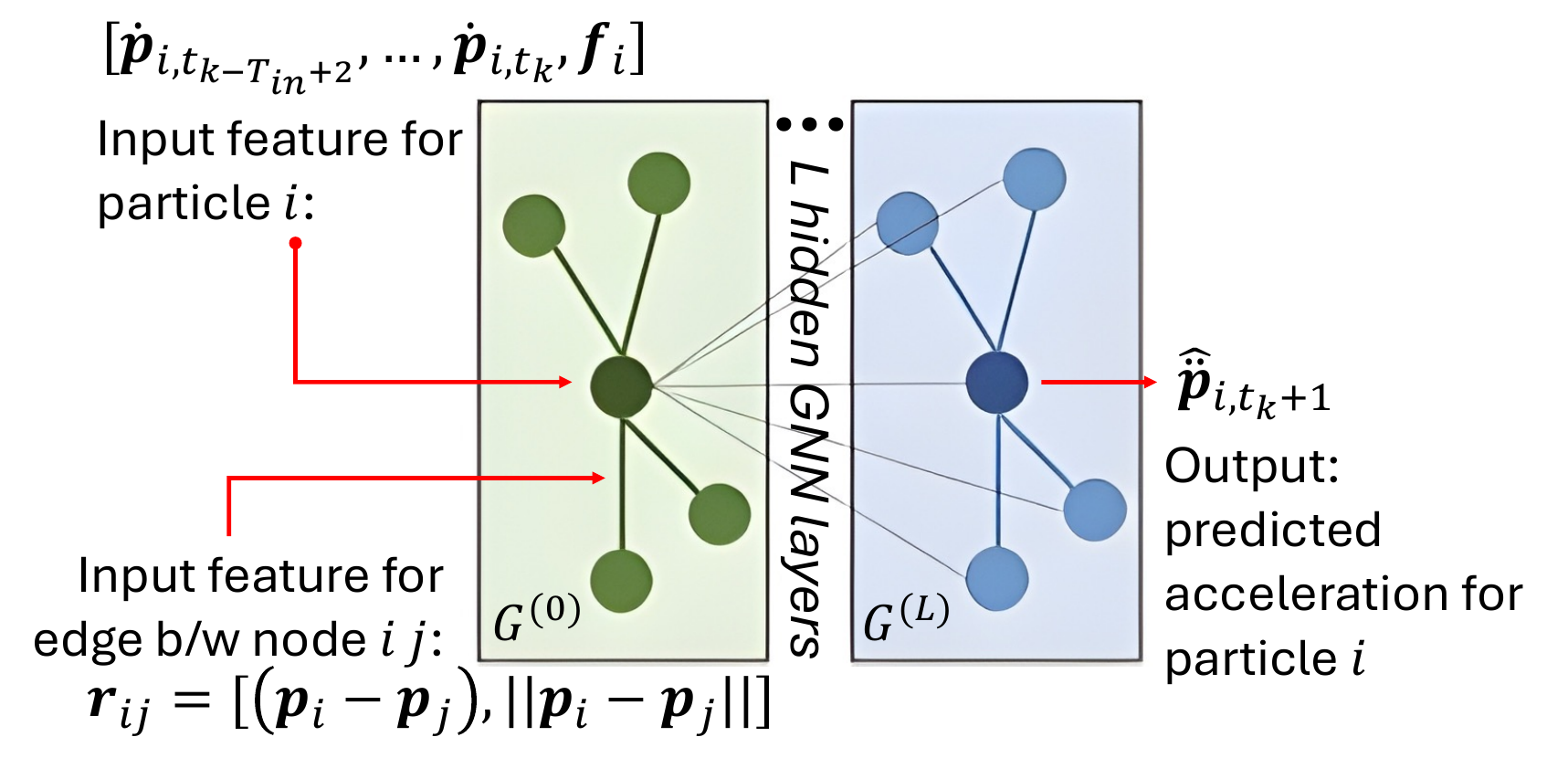}
    \vspace{-1.5em}
    \captionsetup{font=small}
	\caption{GNN as our neural physics simulator.}
    \label{fig:gnn}
    \vspace{-1.5em}
\end{wrapfigure}

\textbf{Input.}
Our neural physics simulator $\simulator_\theta$ takes the input of particle $i$ as:
a sequence of 5 previous velocities (via finite differences from $T_\text{in} = 6$ previous locations),
and features for materials  (e.g., water, sand, rigid, boundary),
i.e.,
$
\bm{x}_{i,t_{k - T_\text{in}}:t_k} =
[\dot{\bm{p}}_{i,t_{k - T_\text{in} + 2}}, \dots, \dot{\bm{p}}_{i, t_k}, \bm{f}_i]$
at time step $t_k$ (Figure~\ref{fig:gnn}).

\textbf{GNN Design.}
We first build the initial graph $G^{(0)}$ by assigning a node to each particle and connecting particles as edges within a fixed ``connectivity radius'' $R$.
The edge embeddings
are learned
from
relative positional displacement and the magnitude
${\bm{r}_{i,j} = [(\bm{p}_i - \bm{p}_j), \norm{\bm{p}_i - \bm{p}_j}]}$.
Our neural physics consists of a stack of $L=10$ GNN layers.
The decoder
predicts the per-particle acceleration, $\ddot{\bm{p}}_i$.
The training loss is the particle-level
$\mathrm{RMSE}_{\ddot{\bm{p}}} \equiv \frac{1}{N} \sum_{i=1}^N \frac{\left\|\hat{\ddot{\bm{p}}}_i-\ddot{\bm{p}}_i\right\|_2}{\left\|\ddot{\bm{p}}_i \right\|_2}$, where $\hat{\ddot{\bm{p}}}_i$ is the predicted acceleration from $\simulator_\theta$.
The future position and velocity are updated using an Euler integrator.
See Appendix~\ref{appendix:gnn_details} for further details.

\section{Methods}
\label{sec:methods}

\begin{figure}[h!]
\vspace{-0.5em}
	\centering
	\includegraphics[width=0.62\textwidth]{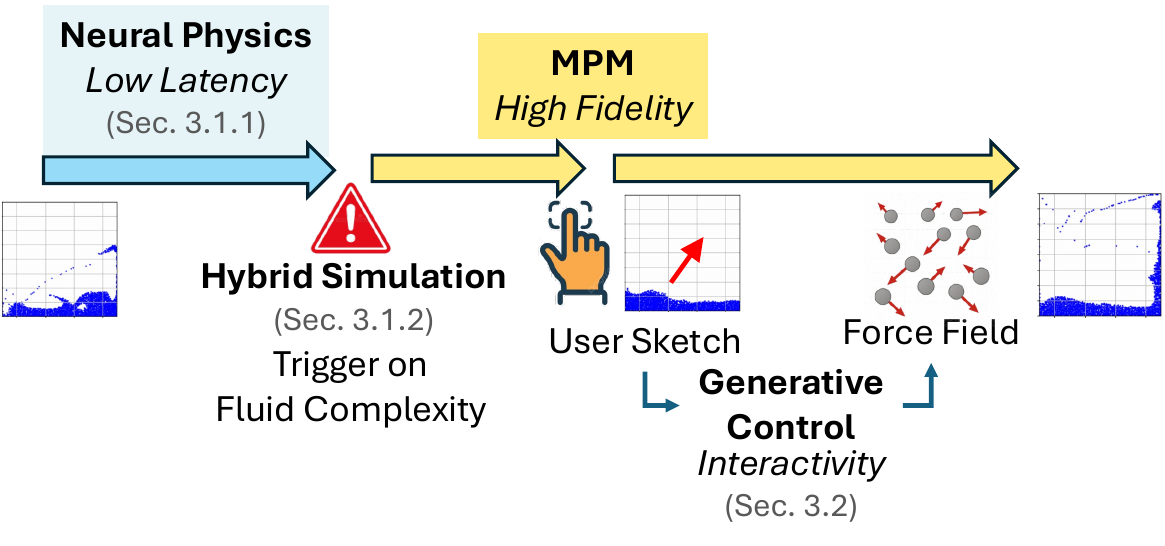}
    \vspace{-1em}
    \captionsetup{font=small}
	\caption{Method Overview.
To achieve real‑time simulations, we cut latency by learning neural physics at a coarse spatiotemporal resolution, while safeguarding fidelity by automatically falling back to an MPM solver when complex fluid phenomena arise (Section~\ref{sec:realtime_hybrid}).
For interactive control, we train a diffusion‑based generative model that infers external force fields directly from user sketches (Section~\ref{sec:interactive_control}).}
    \label{fig:overview}
    \vspace{-1em}
\end{figure}

We aim at real-time fluid simulations (Section~\ref{sec:realtime_hybrid}) with interactive control (Section~\ref{sec:interactive_control}).
Our method is overviewed in Figure~\ref{fig:overview}.

\subsection{Hybrid Real-Time Fluid Simulation}
\label{sec:realtime_hybrid}

\subsubsection{Learning Real-Time Neural Physics at Low Spatiotemporal Resolution}
\label{sec:low_spatiotemporal}

\begin{wrapfigure}{r}{46mm}
\vspace{-1.5em}
	\centering
	\includegraphics[width=0.28\textwidth]{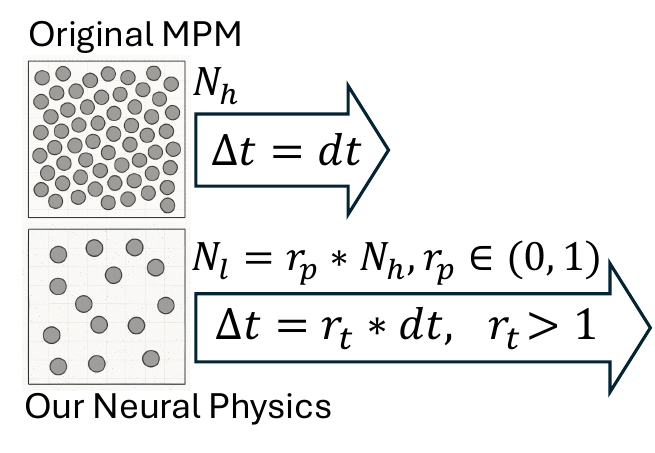}
    \vspace{-0.5em}
    \captionsetup{font=small}
	\caption{Our neural physics accelerates simulations by learning and inferring at low spatial ($N_l$ num. particles) and temporal ($\Delta t$ time steps) resolutions, with downsampling ratios as $r_p$, $r_t$.
    }
    \label{fig:acceleration_spatiotemporal}
    \vspace{-3em}
\end{wrapfigure}

To accelerate the simulation, we train our neural physics at low spatiotemporal resolution.
As shown in Figure~\ref{fig:acceleration_spatiotemporal}, we consider learning the neural physics on simulations with both a downsampled number of particles (ratio $r_p \in (0, 1)$) and also with a larger time step (i.e. coarser temporal discretization rate $r_t \in \mathbb{N}, r_t > 1$).

However, a key pitfall is that once the number of particles is downsampled ($N_h$ particles are merged via clustering into $N_l$, {see Appendix~\ref{appendix:implementation_details}}), we will lose the particle-wise correspondence, i.e., $\hat{\ddot{\bm{p}}}_i$ ($i \in [1, N_l]$) and $\ddot{\bm{p}}_j$ ($j \in [1, N_h]$) cannot align in the particle-level $\mathrm{RMSE}_{\ddot{\bm{p}}}$ (Section~\ref{background:neural_physics}).
As a result, $\mathrm{RMSE}_{\ddot{\bm{p}}}$ can no longer quantify the simulation’s fidelity to the ground truth of the original spatial resolution~\cite{huang2021plasticinelab}.
To mitigate this issue, we use normalized grid-level
$\mathrm{RMSE}_{\tilde{\bm{m}}} \equiv \frac{1}{N} \sum_{i=1}^N \frac{\left\|\hat{\tilde{\bm{m}}}_i-\tilde{\bm{m}}_i\right\|_2}{\left\|\tilde{\bm{m}}_i \right\|_2}$
as the evaluation metric, which essentially quantifies the mass distribution.
$\tilde{\bm{m}}$ is the normalize the grid mass ($\tilde{\bm{m}}_i = \frac{\bm{m}_i}{\sum_{i=1}^N \bm{m}_i}$) converted from particles to the grid via $\mathrm{p2g}$, and $\hat{\tilde{\bm{m}}}$ is the prediction by $\simulator_\theta$. $\tilde{\bm{m}}_i$ and $\hat{\tilde{\bm{m}}}$ share the same grid size but can represent mass distributions from different resolutions (number of particles).
During training, we continue to optimize the surrogate loss $\mathrm{RMSE}_{\ddot{\bm{p}}}$ at the low spatial resolution, thereby avoiding additional $\mathrm{p2g}$ operations.

In Figure~\ref{fig:tradeoff_ablation} (a-c), we can see that by tuning spatiotemporal downsampling ratios $r_p, r_t$, we can improve the trade-off between simulation errors and latency.
Based on this ablation study, we will choose $r_p = 1/1.75$ and $r_t = 2$.
With this configuration, on Water 2D, we can reduce the latency of the original neural physics ($r_p = r_t = 1$) by over {78.8\% (from 1.954ms to 0.4048ms)}.

\subsubsection{Hybrid Simulator with Safeguard}
\label{sec:hybrid_solver}

\begin{figure}[b!]
\vspace{-1em}
	\centering
	\includegraphics[width=1.0\textwidth]{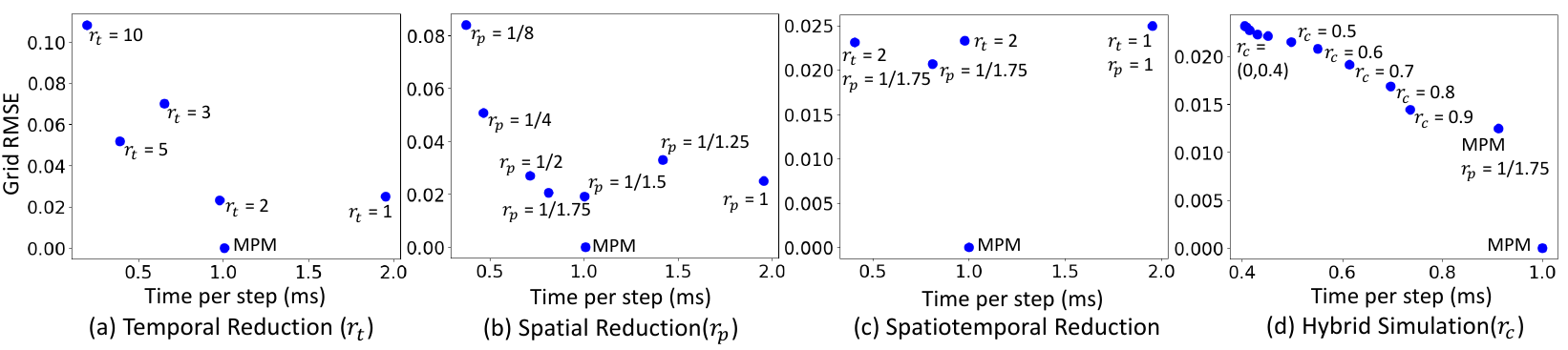}
    \vspace{-1.0em}
    \captionsetup{font=small}
	\caption{Ablation studies of the trade-off between grid-level $\mathrm{RMSE}_{\tilde{\bm{m}}}$ vs. simulation latency. Left to right: temporal reduction $r_t$ (train neural physics with reduced particles $N_l$), spatial reduction $r_p$ (train neural physics with larger time step $\Delta t$), spatiotemporal reduction (combine $r_t = 2$ and $r_p = 1/1.75$), and hybrid with MPM (at $r_p = 1/1.75$) with different thresholds $r_c$.
    Scenario: Water 2D.}
    \label{fig:tradeoff_ablation}
    \vspace{-0.5em}
\end{figure}

Traditional numerical methods like MPM offer high fidelity but are computationally expensive.
Though inferring neural physics at a low spatiotemporal resolution enables significantly faster simulations,
it often comes at the cost of increased simulation errors.
For example, most dots in Figure~\ref{fig:tradeoff_ablation} (a) and (b) are above the original neural physics ($r_p = r_t = 1$) and MPM.

\begin{wrapfigure}{r}{50mm}
\vspace{-1em}
	\centering
	\includegraphics[width=0.35\textwidth]{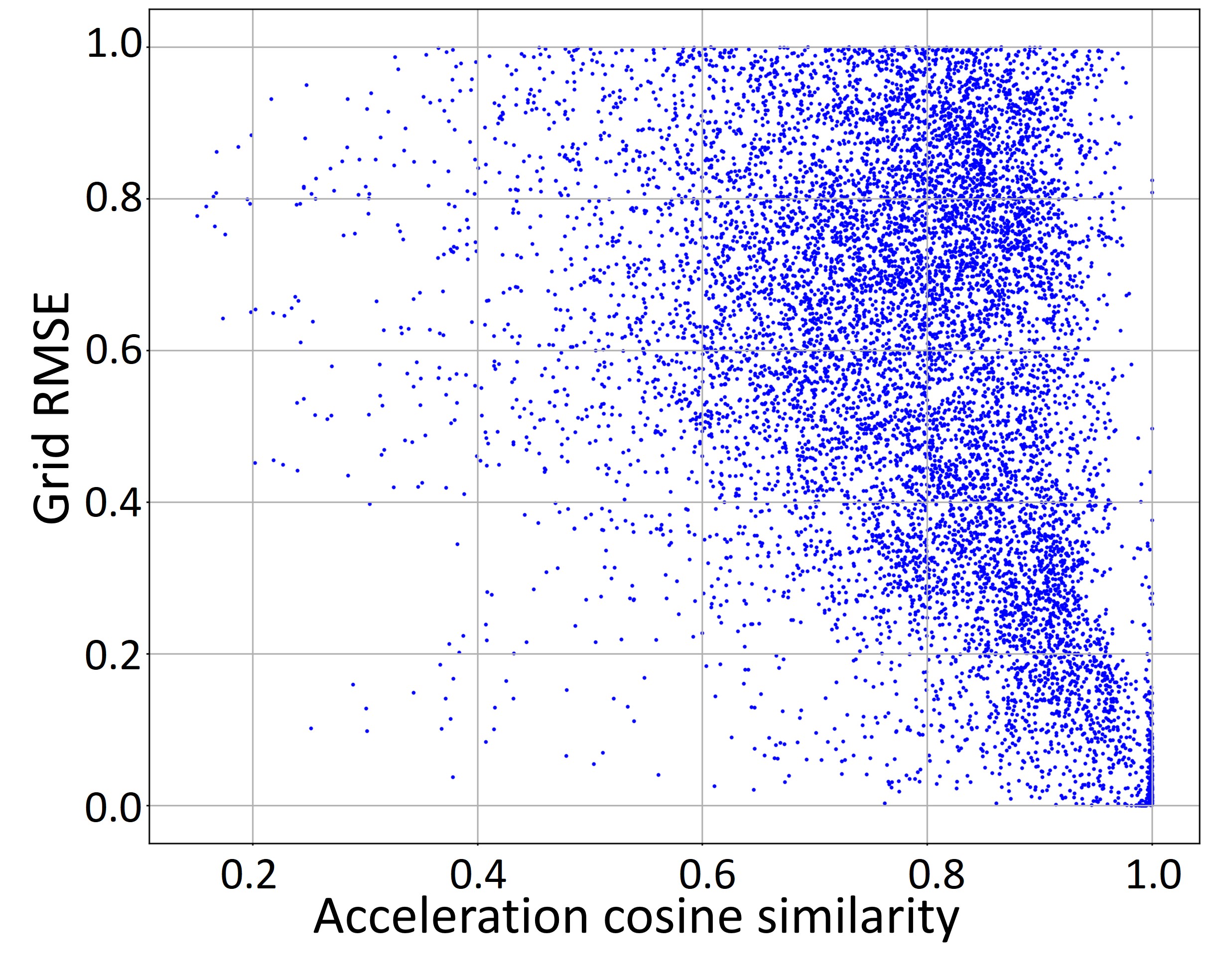}
    \vspace{-0.5em}
    \captionsetup{font=small}
    \caption{Negative correlation between ``cosine similarity of particle accelerations over frames'' vs. ``simulation errors of neural physics''. Scenario: Water 2D. Spearman correlation: -0.3902.}
    \label{fig:cosine_acceleration_error}
    \vspace{-1.5em}
\end{wrapfigure}

To fuse the strengths of both approaches, we make our simulator hybrid.
We primarily leverage neural physics for fast updates, but incorporate a safeguard mechanism to fall back to MPM in challenging scenarios and to empirically ensure simulation quality:
{\small
\begin{equation}
X_{t+1}=\left\{\begin{array}{cl}\text {Neural Physics Update} & \text {if update is ``good''} \\ \text {Fallback to MPM Update} & \text {otherwise. }\end{array}\right.
\end{equation}
}

\paragraph{Fluid Complexity Measures.}

Intuitively, when the current fluid dynamics is simple, the neural physics should generalize well.
In contrast, if the particles behave chaotically, their dynamics become out-of-distribution (OOD) samples that neural physics may be able to generalize.
We thus trigger the fallback condition based on the complexity of the current fluid dynamics being simulated by neural physics.
Moreover, the safeguard should be computationally cheap,
since we need to densely monitor them during the simulation of neural physics.

Specifically, we consider
the cosine similarity of per-particle acceleration over a window of history (window size as $\delta t = 10$ steps by default): $\frac{1}{N} \sum_i^{N} cos(\ddot{\bm{p}}_{i, t-2\delta t:t-\delta t}, \ddot{\bm{p}}_{i, t-\delta t:t})$.
In contrast, we also tried to monitor the divergence of particles' velocity~\cite{gao2025fluidnexus}, which is also used to quantify the quality of incompressible fluid simulations in previous works.
However, its computation is significantly more expensive due to the use of finite difference methods, resulting in increased latency.
We show the negative correlation between this cosine similarity and the neural physics simulation error in Figure~\ref{fig:cosine_acceleration_error}, which indicates that whenever particles' accelerations start diverging, we should fall back to MPM.

\begin{wrapfigure}{r}{75mm}
\vspace{.5em}
	\centering
	\includegraphics[width=0.52\textwidth]{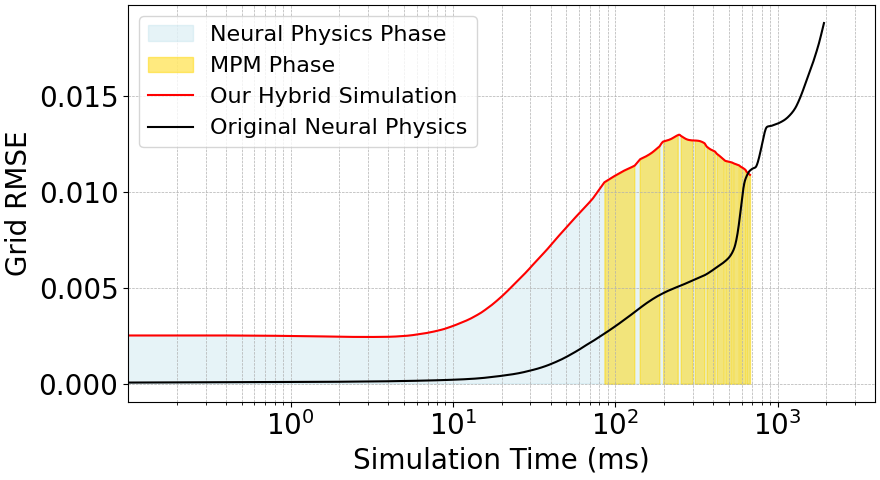}
    \captionsetup{font=small}
    \caption{Error trajectories during simulation (Water 2D). Simulating the \ul{same number of steps ($T=1000$)}, our hybrid solver takes significantly less time (676.4ms) than the original neural physics (1931.1ms), and the final error is also reduced (grid $\mathrm{RMSE}_{\tilde{\bm{m}}}$) (0.0109 vs. 0.0188).
    }
    \label{fig:hybrid_simulation_error_trajectory}
    \vspace{-1em}
\end{wrapfigure}

\paragraph{Triggering MPM by Fluid Complexity.}
With our fluid complexity metric,
we need to trigger the MPM fallback mechanism in principle.
In Table~\ref{table:ablation_hybrid_threshold_nrmse_latency}, we see that when increasing our threshold $r_c$ (i.e. MPM will be more frequently triggered), the simulation fidelity will be corrected by MPM ($\mathrm{RMSE}_{\tilde{\bm{m}}}$ is improved),
and the latency will increase due to heavy computations of MPM.
Thus, we need to choose a threshold $r_c$
such that we can improve our trade-off between $\mathrm{RMSE}_{\tilde{\bm{m}}}$ and latency.
In Figure~\ref{fig:tradeoff_ablation} (d), we
tune this threshold,
and choose {$r_c = 0.8$ to balance the improvements over $\mathrm{RMSE}_{\tilde{\bm{m}}}$ and latency}.

We finalize our hybrid solver using this threshold.
In Figure~\ref{fig:hybrid_simulation_error_trajectory}, we demonstrate trajectories simulated by the original neural physics and our hybrid solver (from the same initial condition, of the same number of steps $T$).
Although the original neural physics ($r_p = r_t = 1$) shows lower rollout errors in the early stage (black curve, due to simulation at high resolution), it quickly accumulates long-term errors. In contrast, after triggering the fallback to MPM (yellow areas), our error is suppressed and we finish the simulation much faster.
Thus, our hybrid solver improves both rollout $\mathrm{RMSE}_{\tilde{\bm{m}}}$
and latency.

\begin{table}[h!]
\centering
\captionsetup{font=small, labelfont=bf, justification=centering}
\caption{Grid $\mathrm{RMSE}_{\tilde{\bm{m}}}$ vs. time per step with hybrid simulations triggered by different thresholds {(Water 2D).}}
\resizebox{1.\textwidth}{!}{
\begin{tabular}{ccccccccccc}
\toprule
Threshold $r_c$ & 0.0 & 0.1 & 0.2 & 0.3 & 0.4 & 0.5 & 0.6 & 0.7 & 0.8 & 0.9 \\
\midrule
Grid $\mathrm{RMSE}_{\tilde{\bm{m}}}$ & 0.0232 &0.0230  & 0.0227 & 0.0223 & 0.0221 & 0.0215 & 0.0208 & 0.0192 & 0.0169 & 0.0144 \\
Time per step (ms) & 0.4048 & 0.4081  & 0.4147 & 0.4301 & 0.4516 & 0.4977 & 0.5509 & 0.6137 & 0.6966 & 0.7356 \\
\bottomrule
\end{tabular}
}
\label{table:ablation_hybrid_threshold_nrmse_latency}
\end{table}

\subsection{Interactive Fluid Control}
\label{sec:interactive_control}

\subsubsection{Use Cases}

\begin{wrapfigure}{r}{75mm}
\vspace{-3em}
	\centering
	\includegraphics[width=0.49\textwidth]{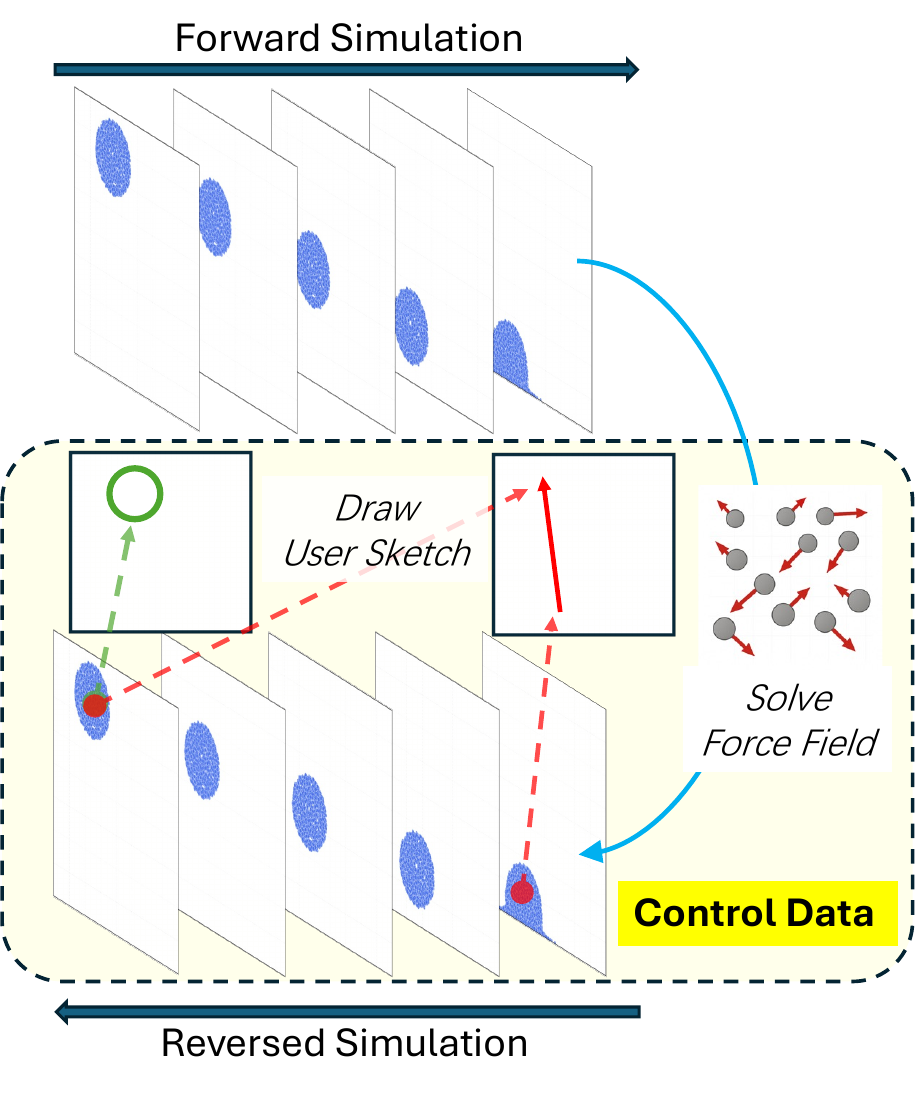}
    \vspace{-1em}
    \captionsetup{font=small}
	\caption{We prepare our training data for generative control via solving external force fields that can reverse a forward simulation. We also prepare user sketches (arrow, ellipse) that depict movements or target shapes of particles (see Appendix~\ref{appendix:implementation_details} for implementation details)}.
    \label{fig:control_data_generation}
    \vspace{-2.5em}
\end{wrapfigure}

Fluid control is essential in computer graphics, where liquid animations convey expressive, story-driven scenes and key visual ideas like splash shapes or motion~\cite{yan2020interactive}.
Manual fluid control produces unnatural effects and forces artists to rely on slow, trial-and-error methods~\cite{pan2013interactive}.
This underscores the need for intuitive tools that let users shape visuals directly, without complex physics.
Yet, achieving the desired appearance of fluid control remains difficult.
Fluid dynamics are intrinsically chaotic and unpredictable.
Setup and tuning of fluid control is tedious and repetitive.
Moreover,
recording real fluid motion is also expensive and hard to customize.

In our paper, we mainly consider the following use case: during a fluid simulation, a user would like to draw a simple sketch and provide it as a control signal, following which the fluid particles should move, as shown in Figure~\ref{fig:control_data_generation} bottom panel.
However, how to \emph{artistically} manipulate fluid particles to follow the user's sketch should be automatically designed by our system.

\subsubsection{Data Generation via Reversed Simulation}
\label{sec:control_data}

The key to making our fluid control possible is to automatically collect training data in principle.
Specifically, we have two highly nontrivial sub-tasks:
1) Design a large number of diverse scenarios of fluid particles with artistic control effects (i.e. fluid particles move along a desired direction or fill a pre-defined shape, in an organized manner, rather than in a chaotic manner);
2) Solve a spatiotemporal external force field that will be applied to the particles, such that the artistic control effect can be fulfilled driven by the composition of gravity, particle interactions, and the proposed force field.

We address these challenges with a reverse simulation strategy.
The core idea is to solve the required force fields that can reverse the fluid dynamics of artistic effects.
We have the following steps:

\paragraph{1) Forward Simulation.}
We randomly simulate a trajectory of fluid dynamics $\bm{X} = (X_1, X_2, \cdots, X_{T_\text{ctr}})$, with different initial conditions (positions or velocities of particles).

\paragraph{2) Reversed Simulation.}
We iteratively solve the required acceleration\footnote{Equivalently, the force field if all particles have the same constant mass} that can restore \emph{positions} of each fluid particle reversely, from $X_{T_\text{ctr}}$ to $X_1$:
\begin{equation}\label{eq:reverse_modeling}
\ddot{\bm{p}}_{t} = \frac{(\bm{p}_{t-1} - \bm{p}_t) - \dot{\bm{p}}_t \cdot \Delta t}{(\Delta t)^2} - \bm{g}
\end{equation}

\paragraph{3) Generation of Control Sketches.}
Finally, based on $\bm{X}$, we generate the user's sketch that depicts the general movements of particles.
We support both directional arrows for movement guidance and one-stroke freehand oval shapes to indicate target regions, as shown in Figure~\ref{fig:control_data_generation}.
See our Appendix~\ref{appendix:implementation_details} for details of implementing freehand arrows and oval shapes.
Note that in 3D scenarios, we use the arrow width to indicate depth~\cite{pan2013interactive}.

For simplicity, we will by default control the fluid particles for 100 MPM steps ($T_\text{ctr} = 100$). That means all our control trajectory will have 100 steps.
While it is possible to employ dynamic neural architectures~\cite{yu2018slimmable} to adaptively adjust the number of MPM steps for this control based on the control complexity, we leave it as a future work.

\subsubsection{Diffusion-based Fluid ControlNet}
\label{sec:controlnet}

Inspired by the recent success of conditioned video generation~\cite{yin2023dragnuwa,wang2024motionctrl,he2024cameractrl,zhao2024motiondirector,wang2024boximator,yang2024direct,xu2024camco,guo2024sparsectrl,wu2024draganything,zhang2024tora},
we choose to train a conditioned diffusion model to control fluid particles.
We by default control the simulations of MPM instead of our neural physics, since the controlled particles may lead to challenging simulations, which are essentially ``OOD'' settings to neural physics. That means, whenever a user provides a control sketch, we will fall back to MPM, and continue the MPM simulation under the control.

\begin{wrapfigure}{r}{70mm}
\vspace{-1em}
	\centering
	\includegraphics[width=0.5\textwidth]{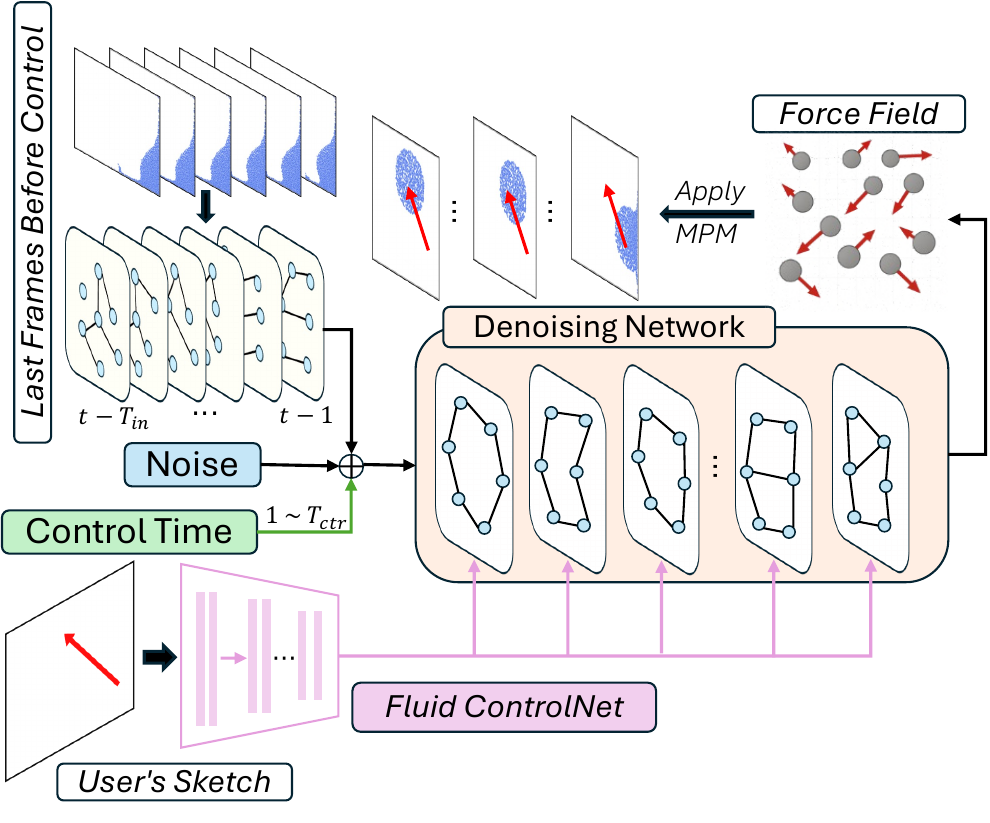}
    \vspace{-1em}
    \captionsetup{font=small}
	\caption{Architecture design of our Fluid ControlNet.}
    \label{fig:controlnet}
    \vspace{-1em}
\end{wrapfigure}

We show our architecture design in Figure~\ref{fig:controlnet}.
Our diffusion-based Fluid ControlNet shares the same backbone and input particle features as our neural physics (Section~\ref{background:neural_physics}).
The output of our Fluid ControlNet is an external force field that will be applied to particles on top of gravity and particle interactions.
Along the MPM simulation, our Fluid ControlNet will unroll the subsequent temporal force fields.
The training target will be the ground truth force fields we simulate in Section~\ref{sec:control_data}.
Parallel to the backbone, we extract the embeddings of the user's sketch input using a convolutional neural network (CNN) and concatenate them with the diffusion timestep embeddings to guide the generation process. We also embed the current control time step into a latent space and integrate it into the initial noise. See Appendix~\ref{appendix:implementation_details} for details of the architecture of our Fluid ControlNet.

\section{Experiments}

\subsection{Settings}

\paragraph{Physical Domains and Simulations.}

\begin{wraptable}{r}{0.33\textwidth}
\vspace{-4.5em}
\centering
\captionsetup{font=small}
\caption{Datasets. $N_h$: Max number of particles at the original spatial resolution. $T$: total time steps. $M$: number of simulation trajectories.}
\vspace{-0.5em}
\resizebox{0.32\textwidth}{!}{%
\begin{tabular}{lccc}
\toprule
Domain & $N_h$ & $T$ & $M$ \\ \midrule
Water (2D) &4k  &1k  &1k  \\
WaterRamps (2D) &3.3k  &600  &1k  \\
Sand (2D) &4k  &320  &1k  \\
SandRamps (2D) &3.3k  &400  &1k  \\
Water (3D) &4k  &800  &1k  \\
Sand (3D) &4k  &350  &1k  \\
Water-Sand (2D) &4k  &500  &1k  \\
\bottomrule
\end{tabular}
}
\label{tab:datasets}
\end{wraptable}

To build our hybrid simulator,
we prepare our own ground truth simulations with the Taichi package~\cite{hu2019taichi,hu2019difftaichi,hu2021quantaichi} on GPUs, with settings closely aligned with~\cite{sanchez2020learning}.
We summarize our scenarios in Table~\ref{tab:datasets}.
We include diverse initial conditions (position, velocity) and numbers of particles.
We fix our grid size as 128 $\times$ 128 for 2D and 64 $\times$ 64 $\times$ 64 for 3D.
We use time step $dt = 2.5$ms in our simulations.

\begin{figure}[t!]
\vspace{-1em}
	\centering
	\includegraphics[width=0.98\textwidth]{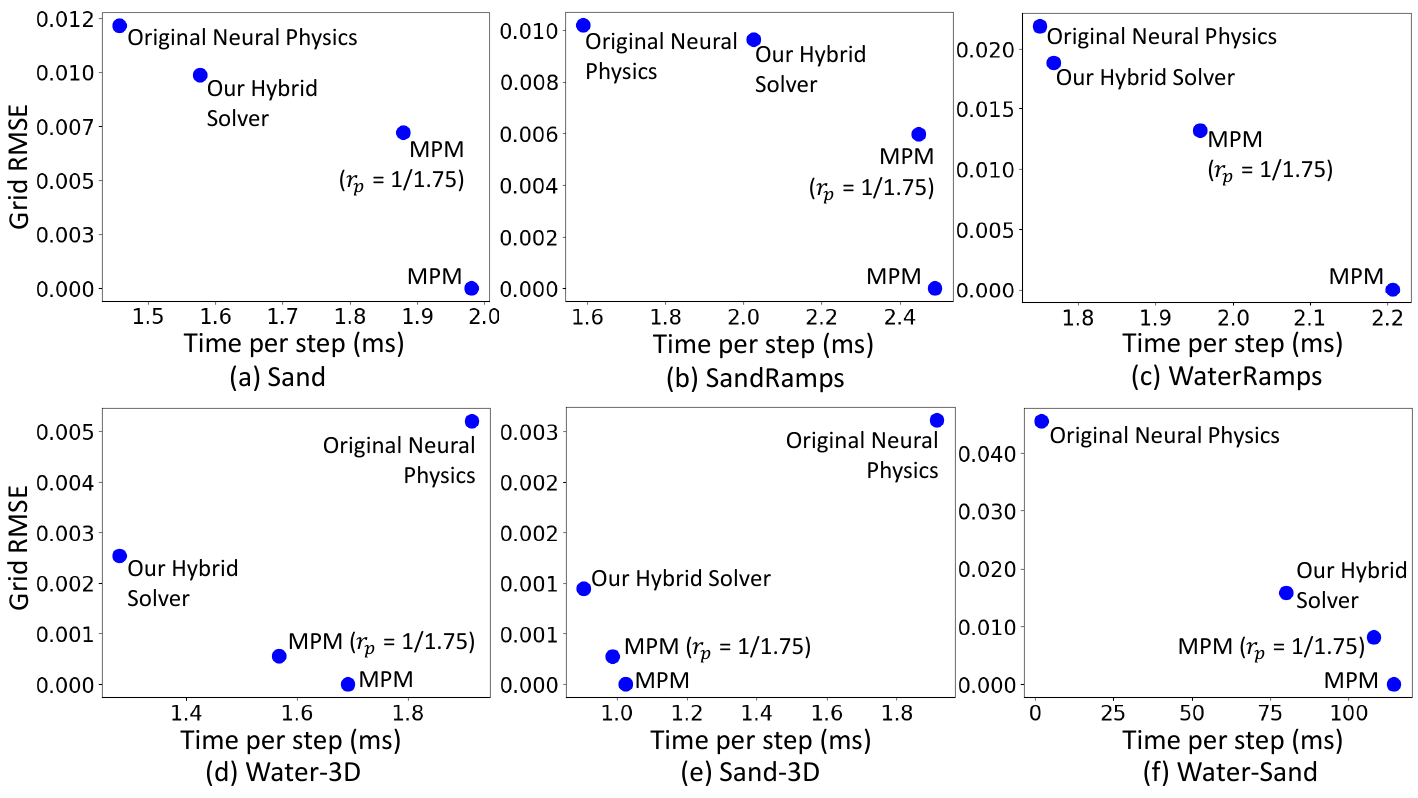}
    \vspace{-0.5em}
    \captionsetup{font=small}
	\caption{Trade-off between simulation error (grid $\mathrm{RMSE}_{\tilde{\bm{m}}}$) and latency, comparing different methods. (a) Sand (2D); (b) SandRamps (2D); (c) WaterRamps (2D); (d) Water (3D); (e) Sand (3D);  (f) Water-Sand (2D).}
    \label{fig:tradeoff_comparison}
    \vspace{-.5em}
\end{figure}

\paragraph{Evaluation.}

{
To report quantitative results, we evaluated our models by computing \textbf{rollout} metrics on held-out test trajectories, drawn from the same distribution of initial conditions used for training.
}
As discussed in Section~\ref{sec:low_spatiotemporal}, we use grid-level $\mathrm{RMSE}_{\tilde{\bm{m}}}$
to compare predictions at lower spatial resolution with the original ground truth.

\subsection{Fluid Simulation Acceleration}
\label{sec:exp_acceleration}

Our hybrid simulator can consistently achieve real-time fluid simulations with preserved simulation fidelity across both 2D and 3D cases.
We show the trade-off between simulation error and latency in Figure~\ref{fig:tradeoff_comparison}, where we compare our hybrid solver with the original neural physics ($r_p = r_t = 1$)~\cite{sanchez2020learning}, MPM~\cite{hu2019taichi,hu2019difftaichi,hu2021quantaichi}, and another MPM that also simulates at low spatial resolution ($r_p = 1/1.75$).
On 2D scenarios, our hybrid solver consistently balances the neural physics and MPM, achieving both reduced simulation latency and preserved simulation errors. For example, on multiple materials (Water-Sand 2D), our hybrid solver can accelerate MPM from 0.114s per frame to 0.08s, with a 29.8\% reduction.
On 3D, the neural physics at $r_p = r_t = 1$ is extremely slow, whereas our hybrid solver improves both latency and errors.
For example, on Sand 3D, we reduce the latency of MPM by 11.8\%, from 1.02 ms to 0.90 ms.

\subsection{Generative Fluid Control}
\label{sec:exp_control}

\begin{figure}
\centering
\includegraphics[width=\textwidth]{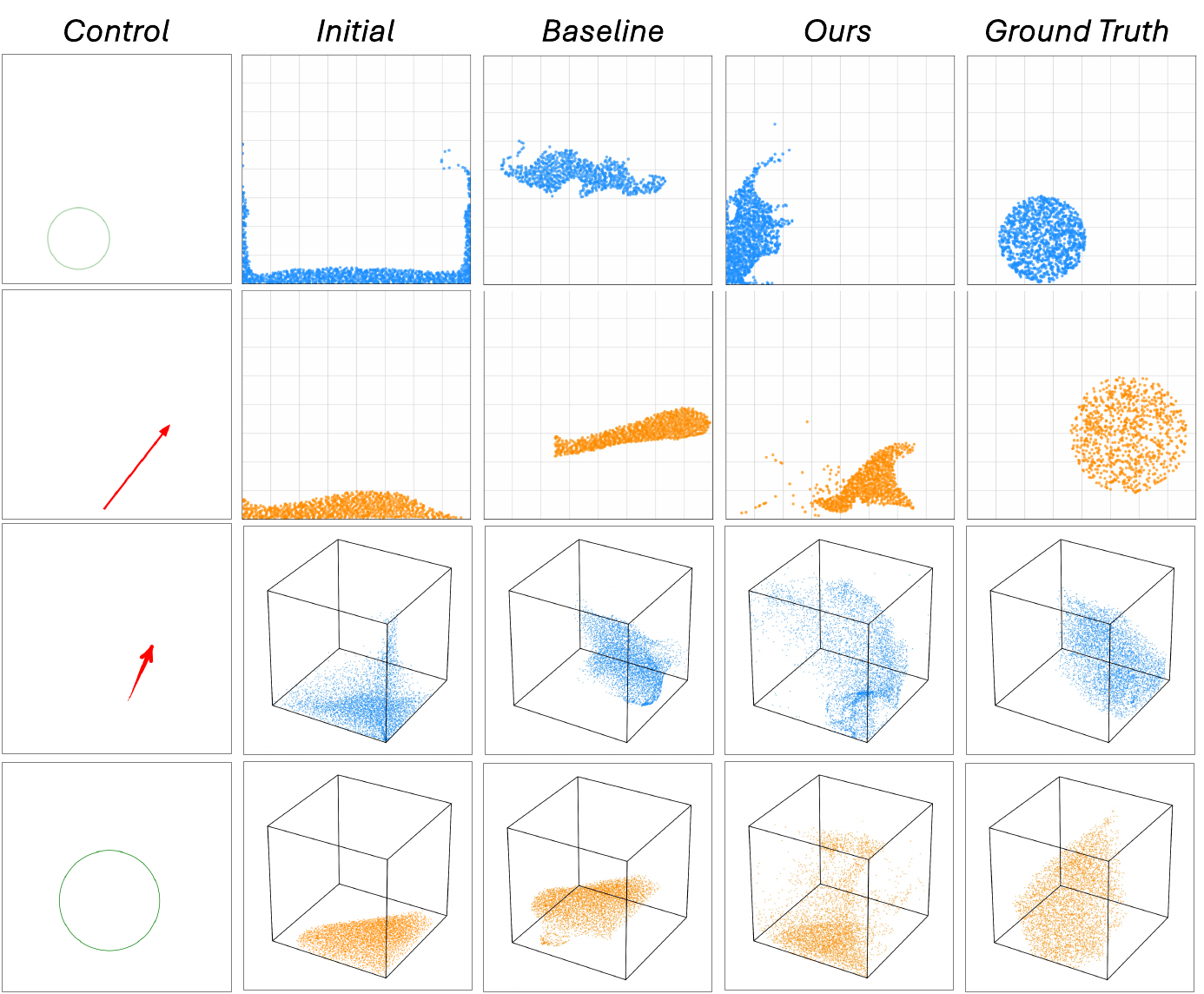}
    \vspace{-0.5em}
    \captionsetup{font=small}
	\caption{Visualization of generative fluid control. Rows from top to bottom:
    Water (2D), Sand (2D),
    Water (3D), Sand (3D).
}
    \label{fig:control_demos}
    \vspace{-.5em}
\end{figure}

We show visualizations of our generative fluid control in Figure~\ref{fig:control_demos}.
We compare with a baseline, where particles are controlled with a spatiotemporal constant force field, with the force magnitude and orientation solved by moving particles from $X_{T_\text{ctr}}$ to $X_1$.

We also quantitatively evaluate the control in Table~\ref{table:control_nrmse}, where we calculate the grid-level $\mathrm{RMSE}_{\tilde{\bm{m}}}$ between the ground truth and the prediction at the last time step, since our main concern is the recovery of the shape of the ground truth at the end of the simulation.
In sum, we can see that our diffusion-based Fluid ControlNet can move particles to better align with the user sketches.

\begin{table}[t!]
\captionsetup{font=small}
\caption{Grid $\mathrm{RMSE}_{\tilde{\bm{m}}}$ between ground truth and predictions at the last time during fluid control.}
\centering
\resizebox{0.58\textwidth}{!}{%
\begin{tabular}{lcccc}
\toprule
Method & Water (2D) & Sand (2D) & Water (3D) & Sand (3D) \\ \midrule
Baseline &0.0908&  0.1151& 0.0019&0.0022    \\
Ours &0.0802&  0.0924&  0.0013& 0.0019  \\ \bottomrule
\end{tabular}
}
\label{table:control_nrmse}
\vspace{-0.5em}
\end{table}

\subsection{Complete Results: Hybrid Simulation + Fluid Control}

\begin{figure}[h!]
\vspace{-0.5em}
	\centering
	\includegraphics[width=0.9\textwidth]{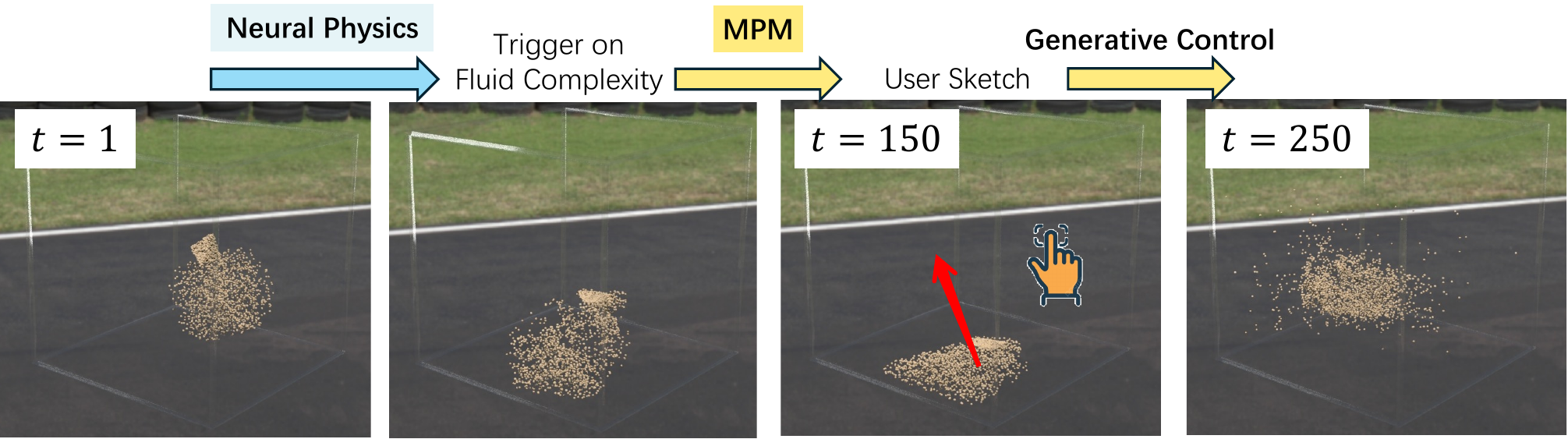}
    \vspace{-0.5em}
    \captionsetup{font=small}
	\caption{Complete results: hybrid simulation + fluid control.
We start the simulation with our neural physics, which is then triggered to MPM. At $t=150$, a user presents the control sketch.
}
    \label{fig:pipeline_results}
    \vspace{-.5em}
\end{figure}

Finally, we present the result from our complete pipeline in Figure~\ref{fig:pipeline_results}.
Particles are first simulated by our hybrid solver, where we start with the neural physics (at low spatiotemporal resolution) and is triggered to MPM once the fluid complex is high.
Then, a user draws a sketch to control, and our diffusion-based Fluid ControlNet takes both this sketch and recent particle states as inputs, and generates external force fields to control particles.

\section{Related Works}

\paragraph{Fluid Modeling and Animation}

Learning-based fluid simulators have progressed from graph-based models to hybrid, physics-informed approaches. DPI-Net~\cite{li2018learning} introduced dynamic interaction graphs with hierarchical message passing to model interactions across particles. This was unified in GNS~\cite{sanchez2020learning,pfaff2020learning,kumar2022gns,kumar2023accelerating}, enabling generalized simulation of fluids, solids, and deformables. Hybrid solvers like MPMNet~\cite{li2023mpmnet} and NeuralMPM~\cite{rochman2024neural} adopt the Material Point Method for scalability. Neural SPH~\cite{toshev2024neural} integrates SPH priors to stabilize rollouts, while NeuroFluid~\cite{guan2022neurofluid} combines learned dynamics and rendering from videos. These advances balance physical accuracy with real-time performance.
In our work, we propose a hybrid approach that combines neural and numerical methods to enable accelerated and high-fidelity fluid simulation.

\paragraph{Fluid Control}

Recent work in fluid control aims to make simulations more intuitive and accessible. Traditional methods using space-time optimization were costly and hard to tune. Yan et al.\cite{yan2020interactive} addressed this with a sketch-based system using conditional GANs to generate liquid splashes. Pan et al.\cite{pan2013interactive} enabled interactive control through sketching and mesh dragging. Chu et al.\cite{chu2021learning} used GANs to infer fluid motion from static fields with semantically controllable features. Schoentgen et al.\cite{schoentgen2020particle} introduced reusable templates for particle-based animations. These approaches shift toward flexible, artist-friendly tools.
We tackle the case where only a freehand sketch is given, and the generative controller is tasked with producing the intended artistic fluid behavior.

\paragraph{Controllable Video Generation}

Controllable video generation has advanced rapidly with diffusion models, especially in disentangling motion control. DragNUWA~\cite{yin2023dragnuwa} enabled trajectory-based editing, while MotionCtrl~\cite{wang2024motionctrl} and Direct-a-Video~\cite{yang2024direct} decoupled camera and object motion. CameraCtrl~\cite{he2024cameractrl} and CamCo~\cite{xu2024camco} refined camera control using geometric cues. MotionDirector~\cite{zhao2024motiondirector} and Boximator~\cite{wang2024boximator} allowed user-customized motion, and SparseCtrl~\cite{guo2024sparsectrl} enabled sparse, entity-level conditioning. Tora~\cite{zhang2024tora} unified text, image, and trajectory inputs for physics-aware generation.
Inspired by these approaches, we leverage forward simulations and compute control forces via reversed simulation.

\section{Conclusion}

In this work, we introduced a novel hybrid neural physics framework that effectively bridges the gap between high-fidelity physical simulation and real-time interactive control.
By combining learned graph-based neural simulators with a fallback to classical MPM solvers, we achieved robust, low-latency fluid dynamics capable of handling complex scenarios without sacrificing accuracy. Additionally, we developed a diffusion-based generative controller trained via revserve modeling, enabling intuitive user interaction through freehand sketches for dynamic fluid control. Extensive experiments across 2D and 3D domains demonstrate that our approach not only accelerates fluid simulations but also provides controllable and physically plausible outcomes.
This hybrid paradigm represents a step forward in making real-time, artist-friendly fluid simulation practical for applications in graphics, design, and virtual environments.

\section{Limitations}
Our current limitations are:
1) The control step $T_\text{ctl}$ is fixed at 100 and is not adaptive to the difficulty of the control scenario;
2) Errors are introduced by the inference of neural physics at low resolution.
The potential solutions are:
1) Training the diffusion-based controller to unroll different numbers of steps to adapt to challenging control scenarios;
2) Training a super-resolution model to correct errors introduced by simulating neural physics at low spatial resolution.
However, addressing these limitations is beyond the scope of this paper, and we plan to study them in our immediate future work.

\bibliographystyle{plain}
\bibliography{references}

\newpage
\appendix

\section{Details of Neural Physics Simulator}
\label{appendix:gnn_details}

\subsection{Particle Simulations as Message-Passing on a Graph}

We denote the state of a particle $i$ at time step $t$ as $\bm{x}_{i,t} \in \mathbb{R}^D$,
and the collective state of $N$ particles as $X_t = [\bm{x}_{1,t}, \dots, \bm{x}_{N,t}] \in \mathbb{R}^{N\times D}$.
Applying physical dynamics over multiple timesteps yields a trajectory of particle states,
$\bm{X}_{t_1:t_{T_\text{in}}} = [X_{t_1}, X_{t_2}, \cdots, X_{t_{T_\text{in}}}] \in \mathbb{R}^{T_\text{in} \times N \times D}$.
In essence, the simulator $\simulator: \mathbb{R}^{T_\text{in} \times N \times D} \rightarrow \mathbb{R}^{N \times d}$ ($d=$ 2 or 3 for 2D/3D) leverages the current $T_\text{in}$ particle states as input to predict their future motion, capturing the underlying dynamics using methods ranging from simple Euler integration to advanced numerical or data-driven techniques.
If a simulator is learnable, it can be represented as $\simulator_\theta$, a parameterized function approximator.
The simulator then iteratively computes future states, such as $\tilde{X}_{t_{T_\text{in}+1}} = \simulator(\tilde{X}_{t_1}, \tilde{X}_{t_2}, \cdots, \tilde{X}_{t_{T_\text{in}}})$, where each newly predicted state is appended to simulate a rollout trajectory over time.

Our learnable simulator $\simulator_\theta$ represents the physical system as interacting particles, where dynamics emerge from exchanges of energy and momentum with neighbors. To ensure robust simulation quality, $\simulator_\theta$ must generalize across diverse interaction patterns and physical scenarios. This particle-based approach naturally maps to message passing on a graph, with particles as nodes and pairwise interactions as edges, making graph neural networks (GNNs) a suitable modeling choice.

\subsection{Details of Graph-based Neural Physics}
\label{sec:experiment:architecture}

Following~\cite{sanchez2020learning},
we implement our neural physics with GNN,
and use standard nearest neighbor algorithms \cite{dong2011efficient,chen2009fast,tang2016visualizing} to construct the graph.

\paragraph{Input.}

In our learnable simulator $\simulator_\theta$, the input state vector for each particle $i$ at time step $t_k$ includes a sequence of 5 previous velocities (via finite differences from $T_\text{in} = 6$ previous locations), and static features representing material properties (e.g., water, sand, rigid, boundary particle).
In practice, only the position vectors $\bm{p}_i$ are stored in our datasets; the velocities $\dot{\bm{p}}_i$ and accelerations $\ddot{\bm{p}}_i$ are computed on the fly using finite differences when needed. Formally, the node feature is defined as
\[
\bm{x}_{i,t_{k - T_\text{in}}:t_k} =
[\dot{\bm{p}}_{i,t_{k - T_\text{in} + 2}}, \dots, \dot{\bm{p}}_{i, t_k}, \bm{f}_i] \in \mathbb{R}^D,
\]
where $\bm{f}_i$ denotes the concatenated material-specific features and scene boundary indicators.
Specifically, the dimension of the encoded node feature vector is $D = 30$ for 2D simulations (5 2-dim velocities by finite differences, i.e., $5\times 2=10$; 4 distances from the boundary; 16-dim embedding for the particle type),
or $D = 37$ for 3D simulations (5 3-dim velocities by finite differences, i.e. $5\times 3=15$; 6 distances from the boundary; 16-dim embedding for the particle type).
See Figure~\ref{fig:gnn} for an illustration.
It is important to note that in our Fluid Controlnet (Section~\ref{sec:controlnet}), the input feature dimension $D$ will increase by 16, where we embed the current control timestep into the latent space with another 2-layer MLP with SiLU activation. 

To obtain more informative edge features $\bm{r}_{i,j}$, we use the relative positional displacement between a pair of adjacent particles $i$ and $j$, along with its magnitude:
\[
\bm{r}_{i,j} = [(\bm{p}_i - \bm{p}_j), \| \bm{p}_i - \bm{p}_j \|].
\]
Edges are added between particles that lie within a predefined \textit{connectivity radius} {$R = 0.015$}, which captures local particle interactions.
$R$ is kept constant for all 2D scenarios.
In different 3D scenarios, a larger radius
can be used to accommodate higher-resolution environments.
Although $R$ is fixed in simulations, edges in the graph are still dynamically updated by comparing the current particle-wise distances to $R$.
For full details of these input and target features, we refer readers to~\cite{sanchez2020learning}.

The ${\encoder: \mathbb{R}^{N \times D} \rightarrow \mathcal{G}}$ embeds particle-based states, it can be formulated as:
${G^{(0)} = (\bm{V}^{(0)}, \bm{E}^{(0)}) = \encoder(\bm{X}, \bm{r}_{i,j})}$
The node embeddings $\bm{V}^{(0)} = \encoder_V(\bm{X})$ are learned functions of the particles' states.
The edge embeddings, $\bm{E}^{(0)}_{i,j} = \encoder_E(\bm{r}_{i,j})$, are learned functions of the pairwise properties of the corresponding particles.
We implement $\encoder_V$ and $\encoder_E$ as multilayer perceptrons (MLP), which encode node features and edge features into the latent vectors, $\bm{V}_i$ and $\bm{E}_{i,j}$, of size 128.

The ${\processor: \mathcal{G} \rightarrow \mathcal{G}}$ computes interactions among nodes through $L$ steps of learned message passing and outputs the final graph, ${G^{(L)} = \processor(G^{(0)})}$. Message passing enables information propagation among particles. Our \processor{} consists of a stack of $L=10$ GNN layers, each using separate (non-shared) MLPs for updating node and edge features, along with residual connections between the input and output latent attributes of both nodes and edges. For the Fluid ControlNet setting, an additional MLP layer is used to encode the diffusion timestep and control image features; see Appendix~\ref{appendix:Fluid_controller} for details.

The ${\decoder: \mathcal{G} \rightarrow \mathbb{R}^{N\times d}}$ extracts dynamics information (of the future state) from the nodes of the final latent graph,
${\hat{X} = \decoder(\bm{V}^{(L)})}$.
Our \decoder{}
is an MLP that outputs accelerations $\ddot{\bm{p}}_i$.
The future position and velocity are updated using an Euler integrator.

All MLPs in \processor have two hidden layers with ReLU, followed by an output layer without activation, with a width of 128.
All MLPs are followed by a LayerNorm~\cite{ba2016layer}.

\section{Implementations}
\label{appendix:implementation_details}

\subsection{Latency Measurements}

\paragraph{Latency of Neural Physics.} 

We utilize the \texttt{TensorRT} library to convert the \texttt{PyTorch} model into an \texttt{ONNX} model to accelerate model inference and align it with the acceleration of MPM on the \texttt{Taichi} kernel.
However, since \texttt{TensorRT} does not support the \texttt{aggregation} operation in GNNs (i.e., aggregating information from edges to adjacent nodes), when measuring the latency, we approximate the time cost of this \texttt{aggregation} operation with a matrix multiplication between an adjacency matrix $A \in \mathbb{R}^{N \times N}$ (where $N$ denotes the number of nodes, i.e. particles),
and node features ($\bm{o}$), such that the aggregation becomes $A \cdot \bm{o}$.
All reported latency measurements are based on the median number of nodes across different scenarios in our test datasets.

\paragraph{Latency of Taichi.}

To enable a fair comparison under MPM simulation setting, we applied a matching latency reduction strategy to the \texttt{Taichi} implementation by skipping non-essential overhead.
Specifically, we excluded the time spent on initializing the MPM state (initial positions and velocities of particles) and the cost of initializing the \texttt{Taichi} kernel at the beginning of the simulation.
As a result, our comparison focuses solely on the runtime per simulation step after the \texttt{CUDA} or \texttt{Taichi} kernel has been initialized.

\subsection{Design of Fluid ControlNet}
\label{appendix:Fluid_controller}
In our Fluid ControlNet, the control signal $\mathcal{C} \in \mathbb{R}^{H\times W \times 3}$ is encoded using our Fluid ControlNet. The encoded embedding is then injected into the graph-based diffusion model to guide the generation of the external field of accelerations.
The Fluid ControlNet consists of 8 convolutional layers and 3 downsampling operations.
It extracts multi-scale features from the control signal $\mathcal{C}$, projects each scale to a different dimensional space, and then concatenates the projected features into control embedding representations of dimension size 44.
The resulting embedding is then integrated into the \processor{} module of the graph-based diffusion model. 
Notably, to better condition the diffusion process on the control signal, we draw inspiration from DiT~\cite{peebles2023scalable} and concatenate the embedding of the control signal to the diffusion time step embedding. This design choice ensures that the control condition is effectively incorporated at each diffusion step, thereby generating high-fidelity acceleration fields that can align fluid particles to the target motion or shape.

\subsection{Training}
Following~\cite{sanchez2020learning}, we normalize the input velocity to the GNNs, and apply random noises to input positions ($\bm{p}_{t_1:t_{T_\text{in}}}$) during training.
For both neural physics and Fluid ControlNet,
we train with the Adam optimizer and a learning rate at $1\times10^{-4}$ with exponential decay.
Our training batch size is 1, and we train for 2 million gradient descent steps.

\begin{table}[h!]
\centering
\captionsetup{font=small, labelfont=bf, justification=centering}
\caption{Training Costs (GPU hours)
across different scenarios.}
\resizebox{0.78\textwidth}{!}{
\begin{tabular}{ccccc}
\toprule
GPU Hours & Water (2D) & Sand (2D) & Water (3D) & Sand (3D) \\ \midrule
Neural Physics (Section~\ref{sec:realtime_hybrid})  &17.27h  &17.94h  &19.71h  &19.67h  \\
Fluid ControlNet (Section~\ref{sec:interactive_control})  &  69.87h 
& 76.36h &  184.12h & 151.03h  \\ \bottomrule
\end{tabular}
}
\label{table:training_costs}
\end{table}

We include our training costs in Table~\ref{table:training_costs}.
Neural physics requires approximately one day on a single NVIDIA 4090 GPU.
For the Fluid ControlNet, training takes around three days for 2D scenarios on a single NVIDIA 4090 and six days for 3D scenarios on a single NVIDIA A40.
We train both neural physics and Fluid ControlNet with the
particle-level RMSE loss on predicted accelerations
$\mathrm{RMSE}_{\ddot{\bm{p}}} \equiv \frac{1}{N} \sum_{i=1}^N \frac{\left\|\hat{\ddot{\bm{p}}}_i-\ddot{\bm{p}}_i\right\|_2}{\left\|\ddot{\bm{p}}_i \right\|_2}$, which was defined in Section~\ref{background:neural_physics}.

\subsection{Generating Users' Freehand Sketches (Arrows and Oval Shapes)}
Arrows are computed by connecting the centroid ($\bar{\bm{p}} = \frac{1}{N}\sum_{i=1}^N \bm{p}_i$) of fluid particles at $t=1$ ($\bar{\bm{p}}_1$) and $t=T_\text{ctr}$ ($\bar{\bm{p}}_{T_\text{ctr}}$).
Based on the mean displacement vector $\Delta\bar{\bm{p}} = \bar{\bm{p}}_1 - \bar{\bm{p}}_{T_\text{ctr}}$,
we derive the arrow length $\|\Delta\bar{\bm{p}}\|$ and orientation $\theta = \tan^{-1}(\Delta\bar{\bm{p}}_y / \Delta\bar{\bm{p}}_x)$.
In 3D, we use the arrow width to indicate depth~\cite{pan2013interactive}.
A multi-segment arrow with varying line width is implemented as $n$-segment polyline with width modulation, where each segment's width $w_i$ ($i \in [1, n]$) is
$w_i = w_{\min} + (w_{\max}-w_{\min})\cdot\frac{\Delta\bar{\bm{p}}_{z,i}-\Delta\bar{\bm{p}}_{z,\min}}{\Delta\bar{\bm{p}}_{z,\max}-\Delta\bar{\bm{p}}_{z,\min}}$.
The arrowhead adopts perspective-correct scaling.

For 2D oval sketches, shapes of particles at $t=T_\text{ctr}$ are represented as elliptical outlines centered at $\bar{\bm{p}}_{T_\text{ctr}}$,
with radii corresponding to $\pm2\bm{\sigma}$, where $\bm{\sigma}$ is the standard deviation of particle positions along each principal axis.
This statistically-grounded ellipse captures approximately 95\% particles' positions while being visually simple.
Meanwhile, 2D oval-shaped control sketches can indeed be ambiguous in 3D, since it is infeasible to depict 3D volumes with a simple one-stroke 2D sketch.

\subsection{Enforcing Smoothness on Target Accelerations}

\begin{figure}[h!]
\vspace{-0.5em}
	\centering
	\includegraphics[width=0.49\textwidth]{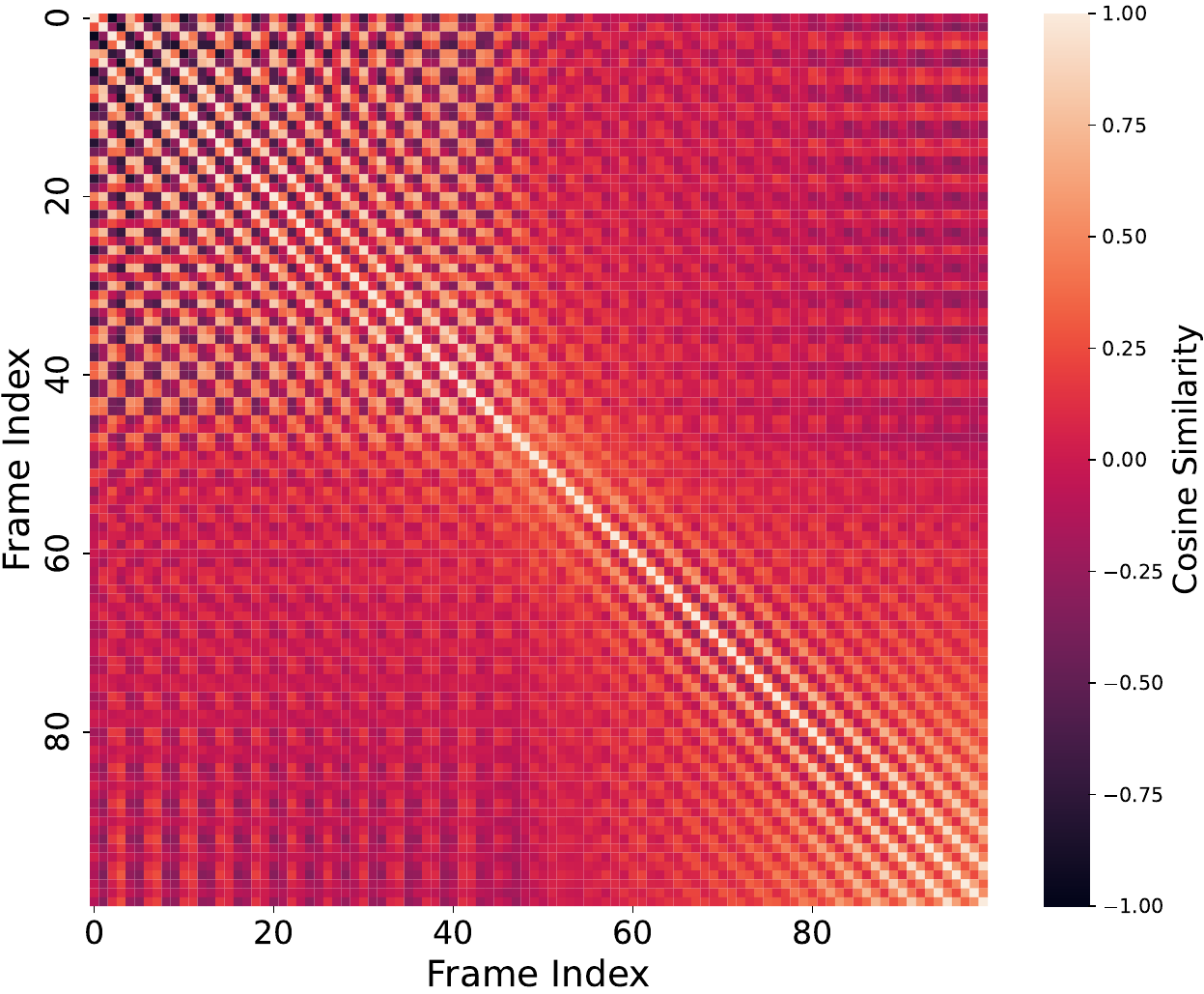}
	\includegraphics[width=0.49\textwidth]{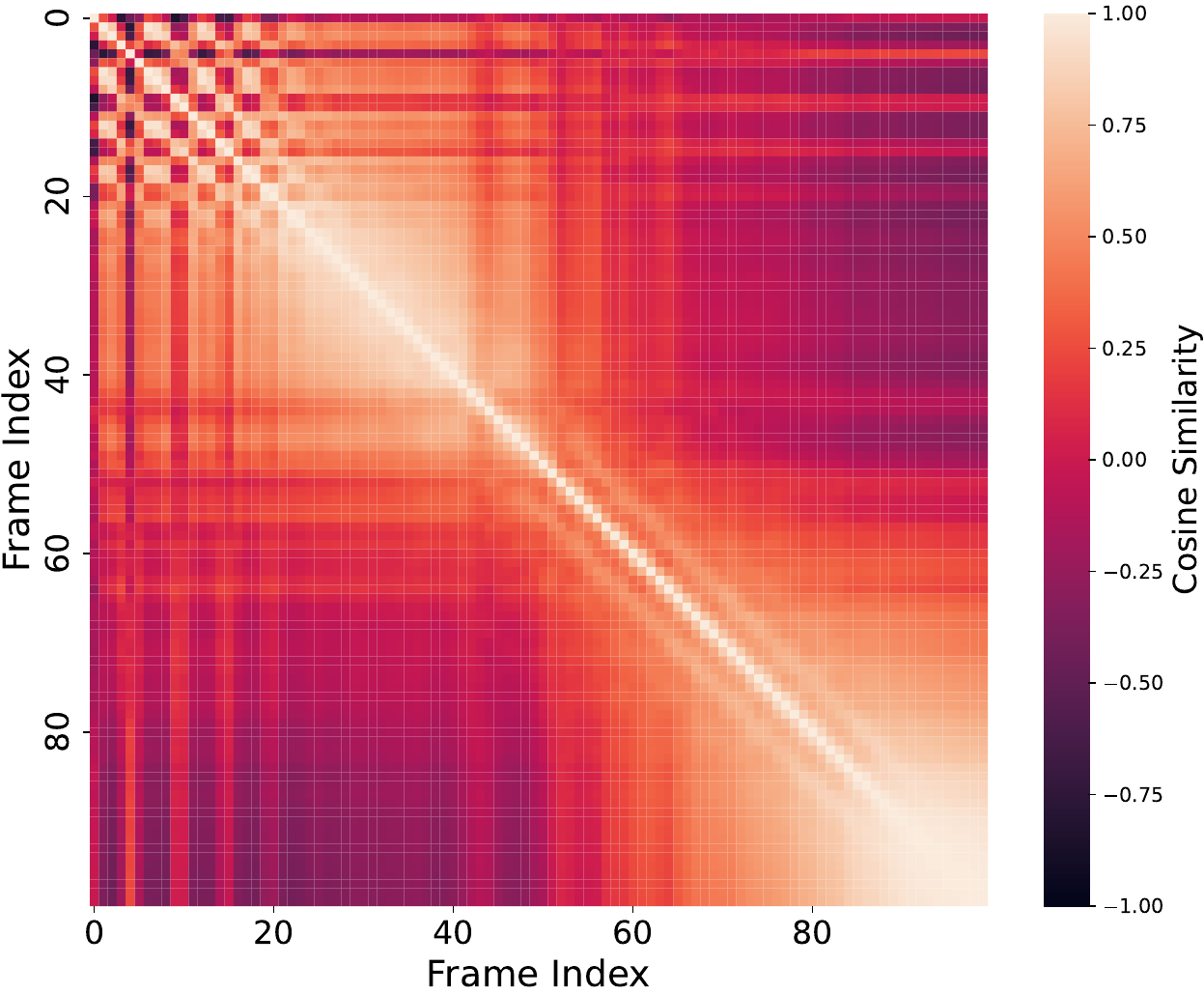}
    \vspace{-0.5em}
    \captionsetup{font=small}
    \caption{Step-wise correlations of ground-truth accelerations for fluid control. Left: before enforcing smoothness; Right: after enforcing smoothness.}
    \label{fig:acceleration_correlation_control}
    \vspace{-1em}
\end{figure}

We observe that ground truth accelerations solved by Equation~\ref{eq:reverse_modeling} are typically complicated (see the temporal-wise cosine similarity in Figure~\ref{fig:acceleration_correlation_control}), which will be challenging to learn.
We thus further enforce a certain level of smoothness of the acceleration across temporal steps:
\begin{equation}
\label{eq:control_acceleration_smoothness}
\ddot{\bm{p}}_{t, \text{smooth}} = \ddot{\bm{p}}_t - \lambda \cdot \exp \left( -\beta \cdot \frac{\ddot{\bm{p}}_t \cdot \ddot{\bm{p}}_{t+1}}{\|\ddot{\bm{p}}_t\| \cdot \|\ddot{\bm{p}}_{t+1}\|} \right) \cdot (\ddot{\bm{p}}_t - \ddot{\bm{p}}_{t+1})
\end{equation}
Essentially, Equation~\ref{eq:control_acceleration_smoothness} enforces decoupled smoothness over the magnitude and the orientation of accelerations over temporal steps.
We choose $\lambda = 0.1$ and $\beta = 2$ in our work.

\section{More Results}

\subsection{Grid $\mathrm{RMSE}_{\tilde{\bm{m}}}$ of Fluid Simulations over Random Seeds}
To ensure a fairer comparison, we conducted experiments using three different random seeds.
The results, as shown in Table~\ref{table:grid_rmse_random_seeds}, demonstrate that our hybrid solver consistently outperforms the original neural physics across all datasets.

\begin{table}[h!]
\centering
\captionsetup{font=small, labelfont=bf, justification=centering}
\caption{Grid $\mathrm{RMSE}_{\tilde{\bm{m}}}$ of fluid simulations on different scenarios, over three random runs.}
\setlength{\tabcolsep}{3pt}
\resizebox{1.\textwidth}{!}{
\begin{tabular}{cccccccc}
\toprule
Grid $\mathrm{RMSE}_{\tilde{\bm{m}}}$ & Water (2D) & Sand (2D) & SandRamps (2D) & WaterRamps (2D) & Water (3D) & Sand (3D) & Water-Sand (2D) \\ \midrule
Neural Physics &0.0263 (1.15e-6)  &0.0125 (2.59e-7) &0.0101 (3.23e-8)  &0.0229 (2.09e-6)  &0.0048 (6.58e-7)  &0.0025 (2.11e-8)  &0.0441 (3.51e-6)  \\
Our Hybrid Solver &0.0186 (8.17e-6)  &0.0116 (6.88e-8)  &0.0096 (1.00e-9)  &0.0171 (3.16e-6)  &0.0022 (1.77e-8)  &0.0013 (1.08e-7)  &0.0149 (2.38e-6)  \\ \bottomrule
\end{tabular}
}
\label{table:grid_rmse_random_seeds}
\end{table}

\subsection{More Visualizations}

\begin{figure}[h!]
\vspace{-0.5em}
	\centering
	\includegraphics[width=0.13\textwidth]{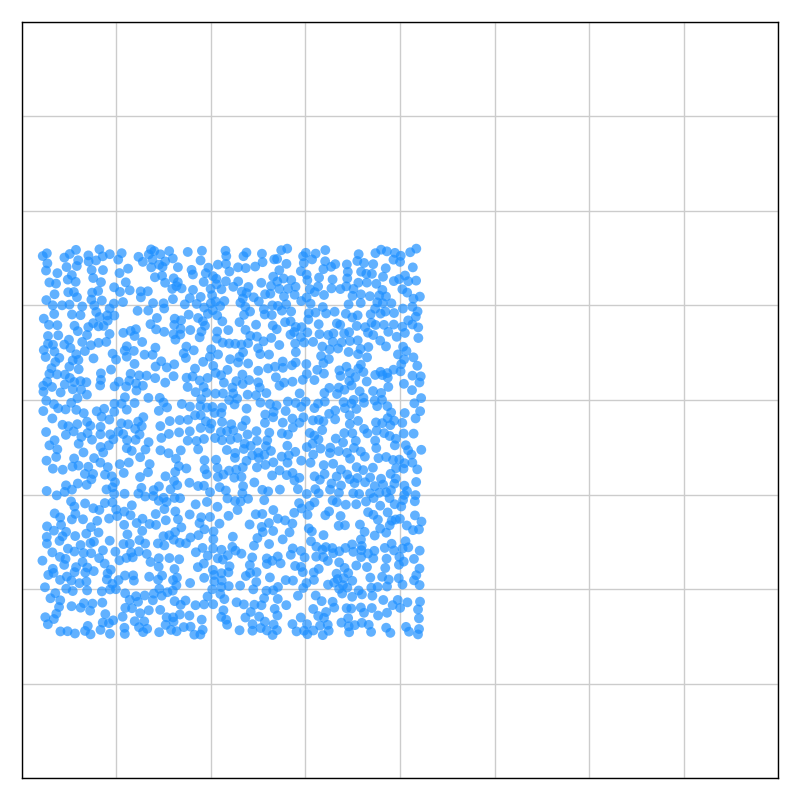}
	\includegraphics[width=0.13\textwidth]{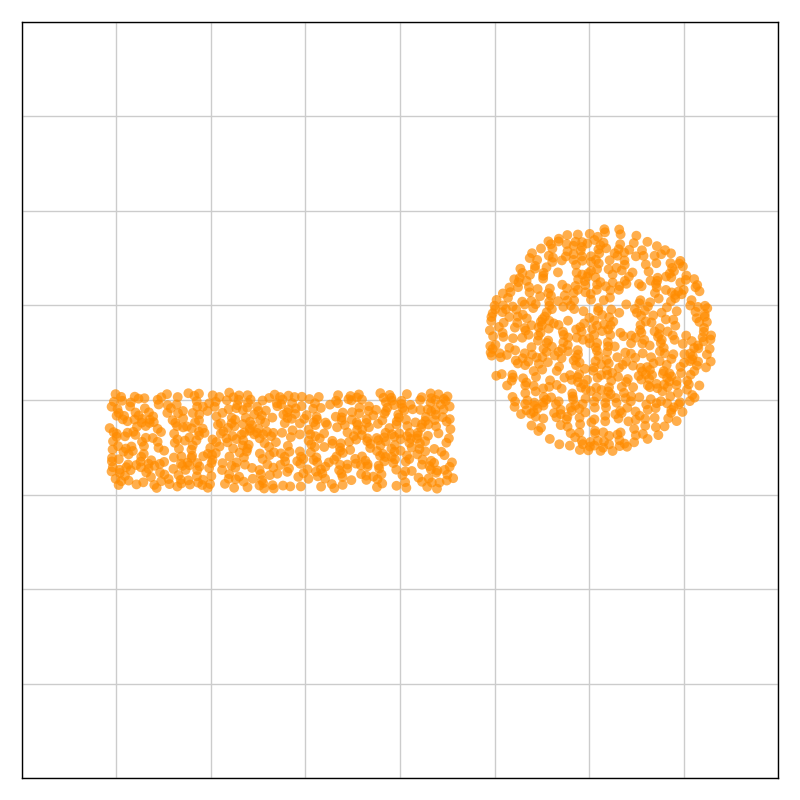}
	\includegraphics[width=0.13\textwidth]{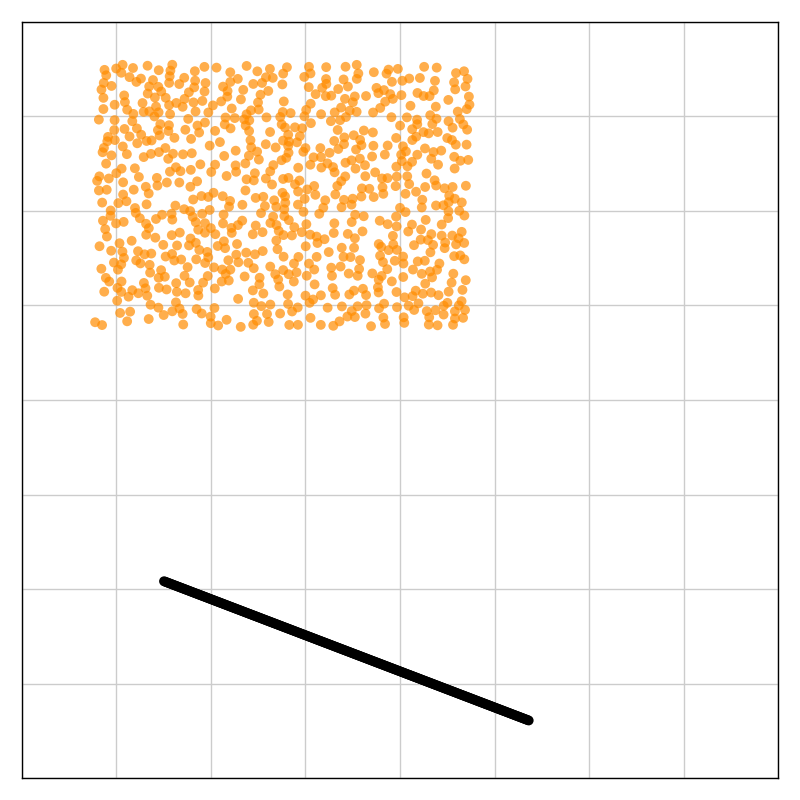}
	\includegraphics[width=0.13\textwidth]{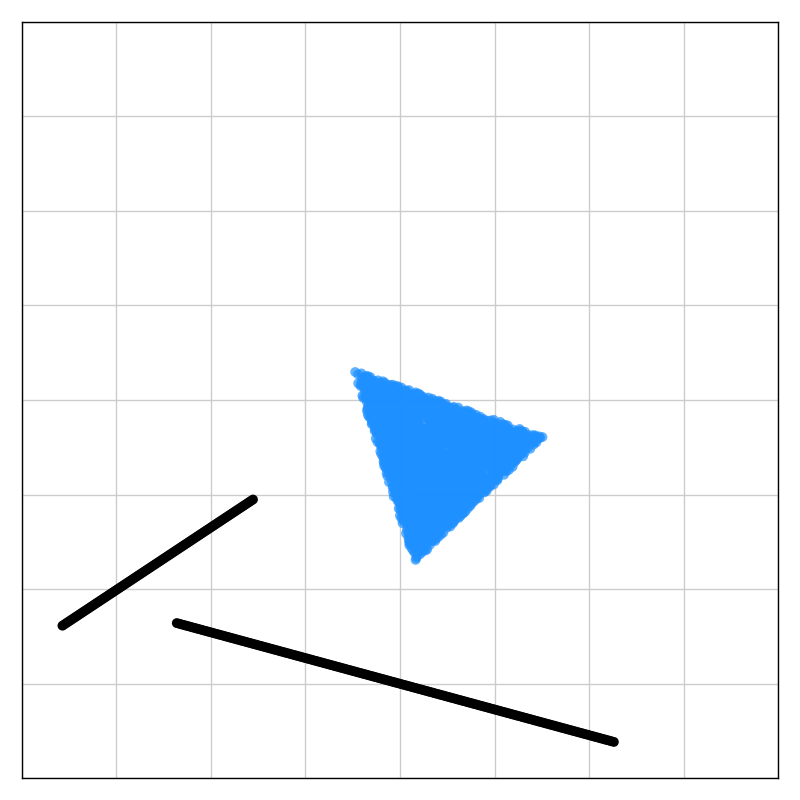}
	\includegraphics[width=0.13\textwidth]{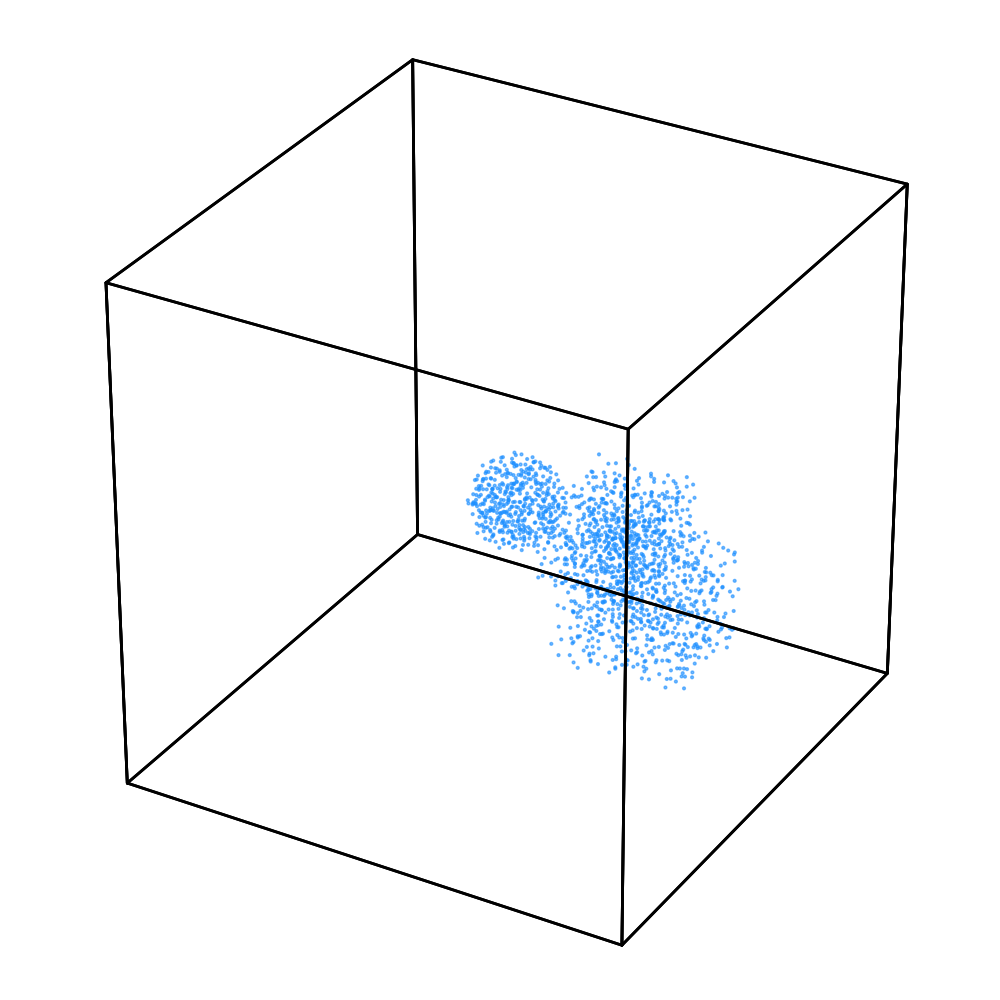}
	\includegraphics[width=0.13\textwidth]{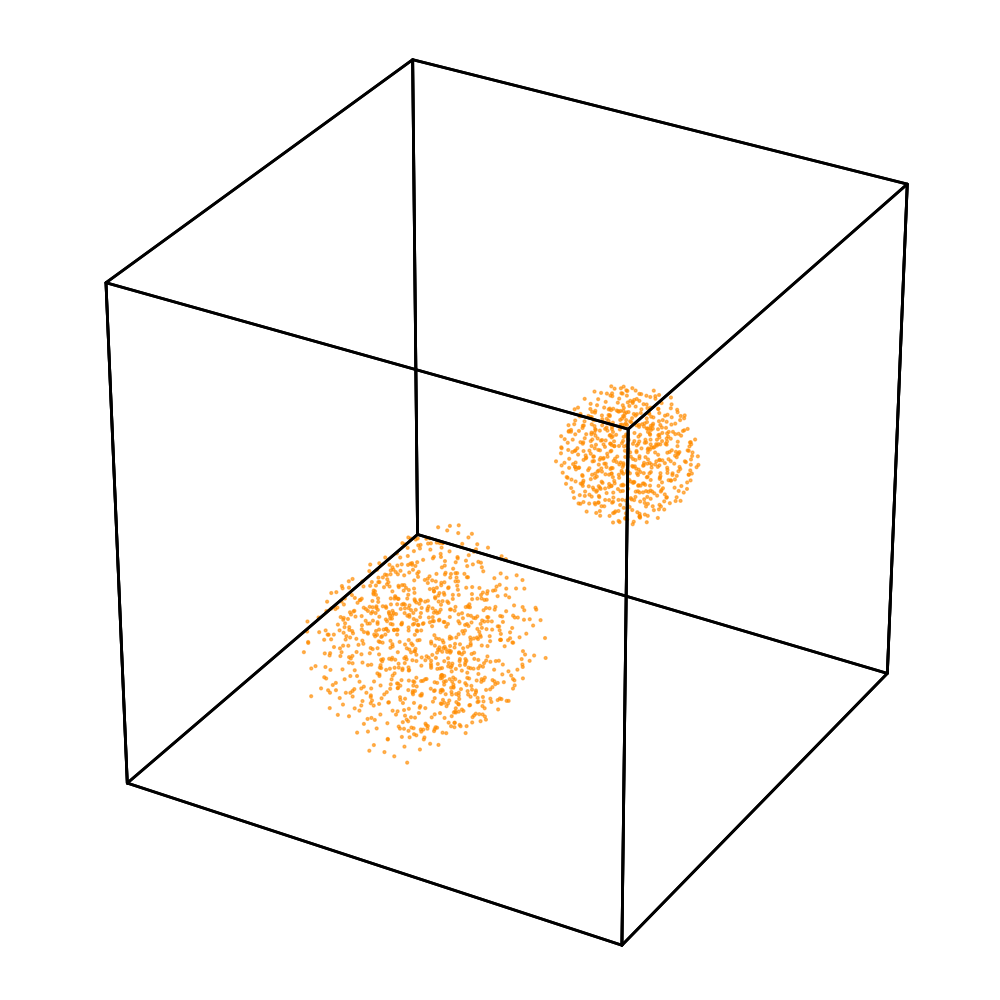}
	\includegraphics[width=0.13\textwidth]{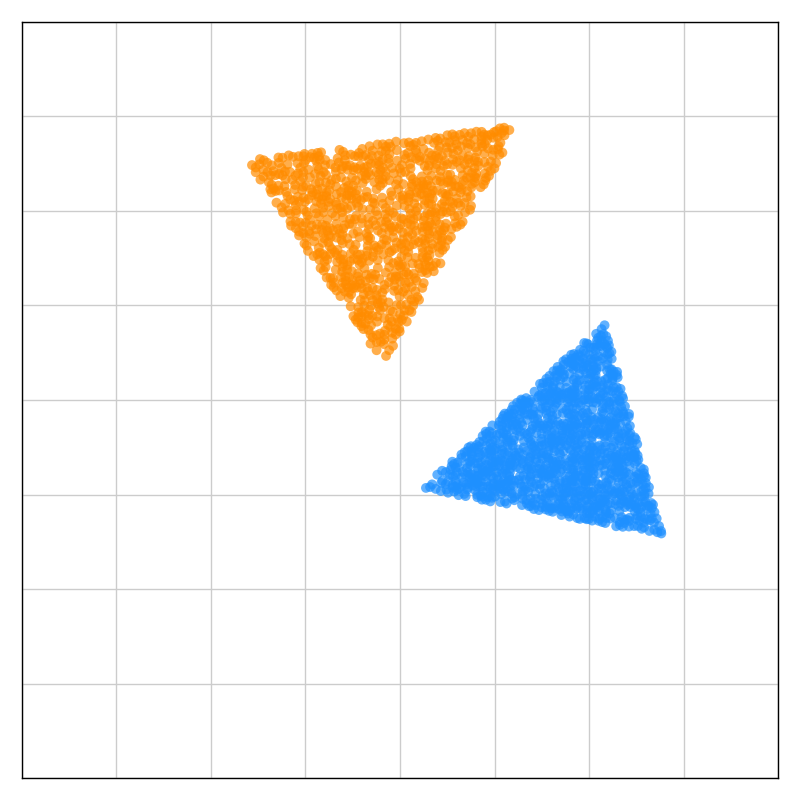}
	\includegraphics[width=0.13\textwidth]{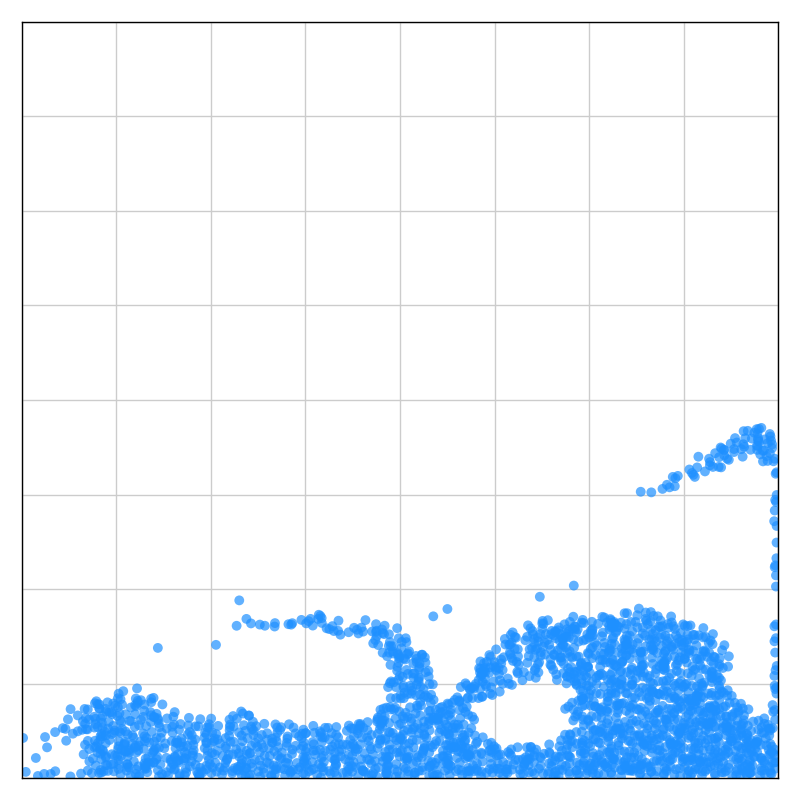}
	\includegraphics[width=0.13\textwidth]{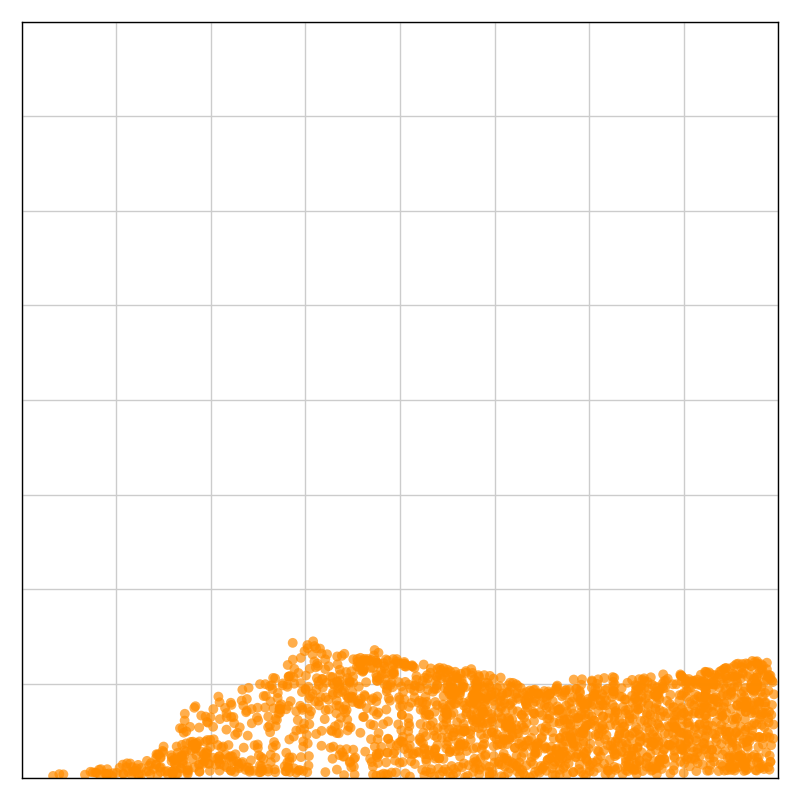}
	\includegraphics[width=0.13\textwidth]{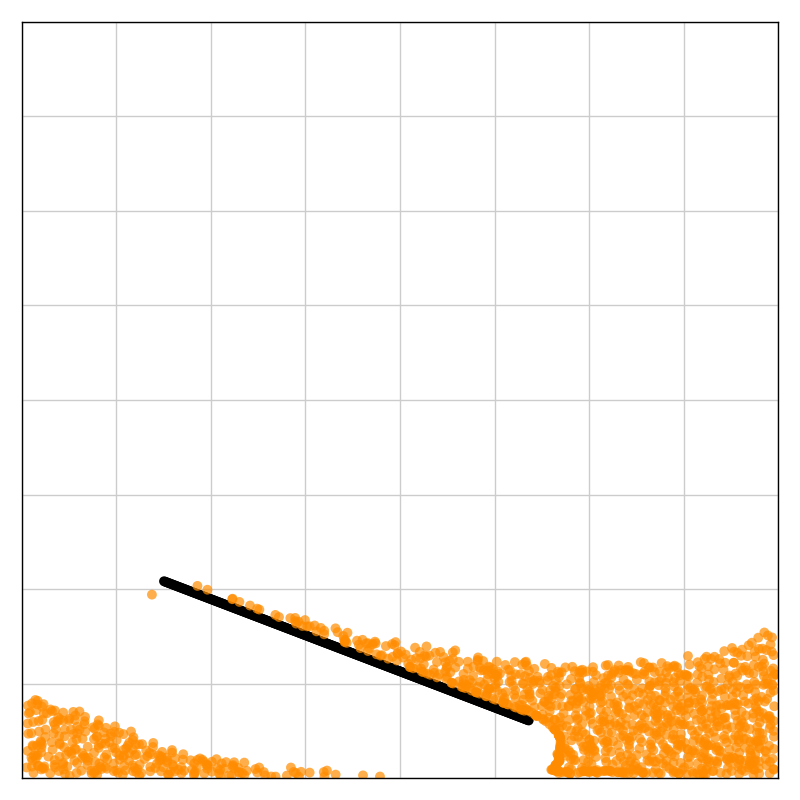}
	\includegraphics[width=0.13\textwidth]{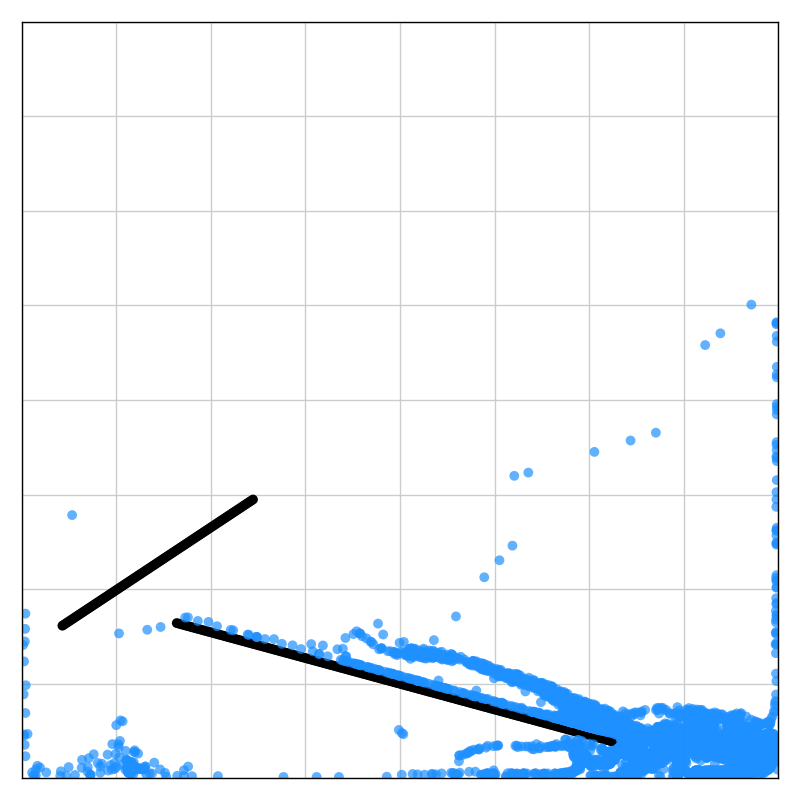}
	\includegraphics[width=0.13\textwidth]{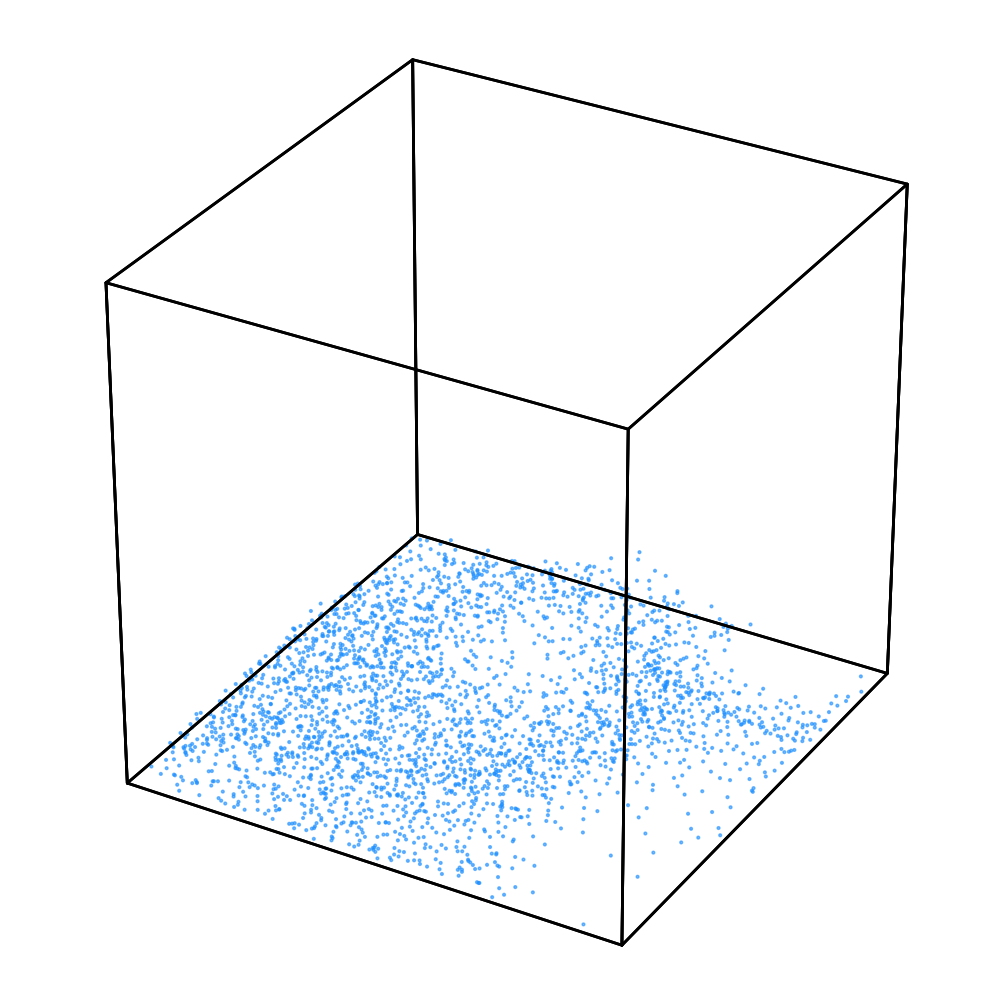}
	\includegraphics[width=0.13\textwidth]{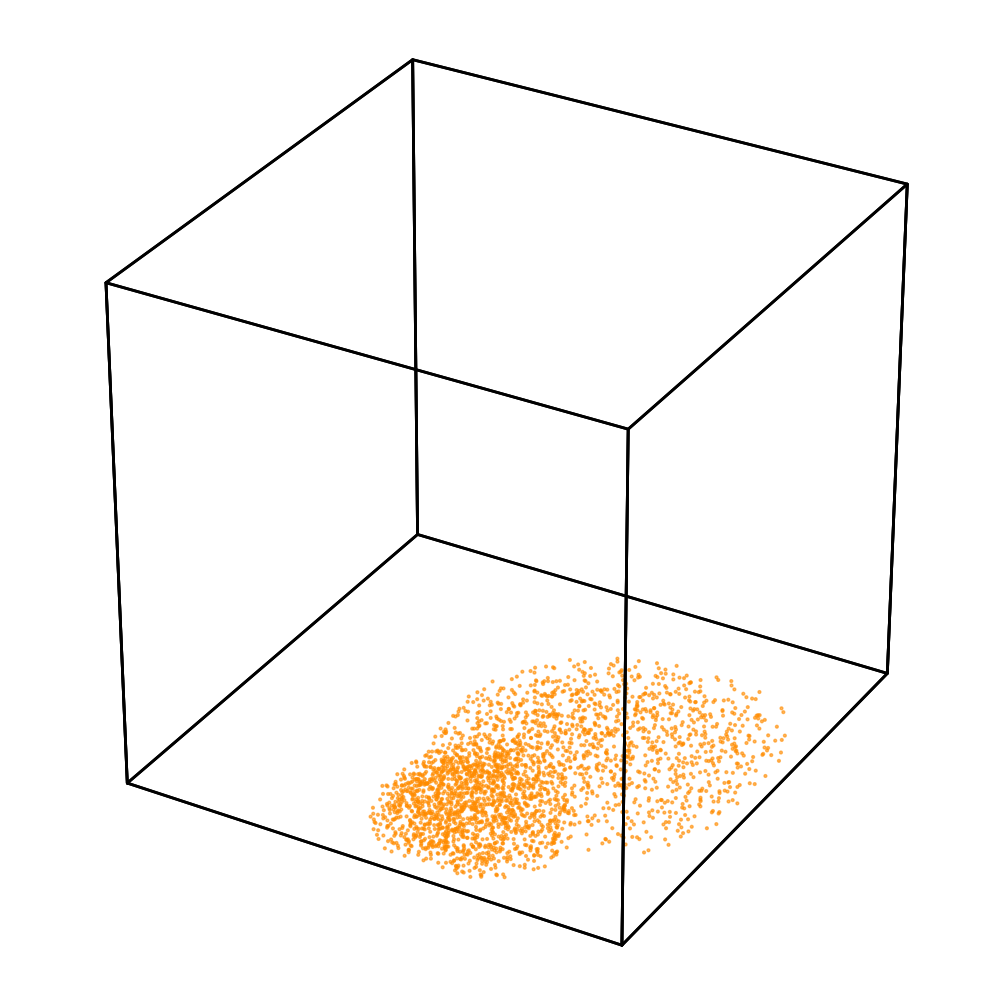}
	\includegraphics[width=0.13\textwidth]{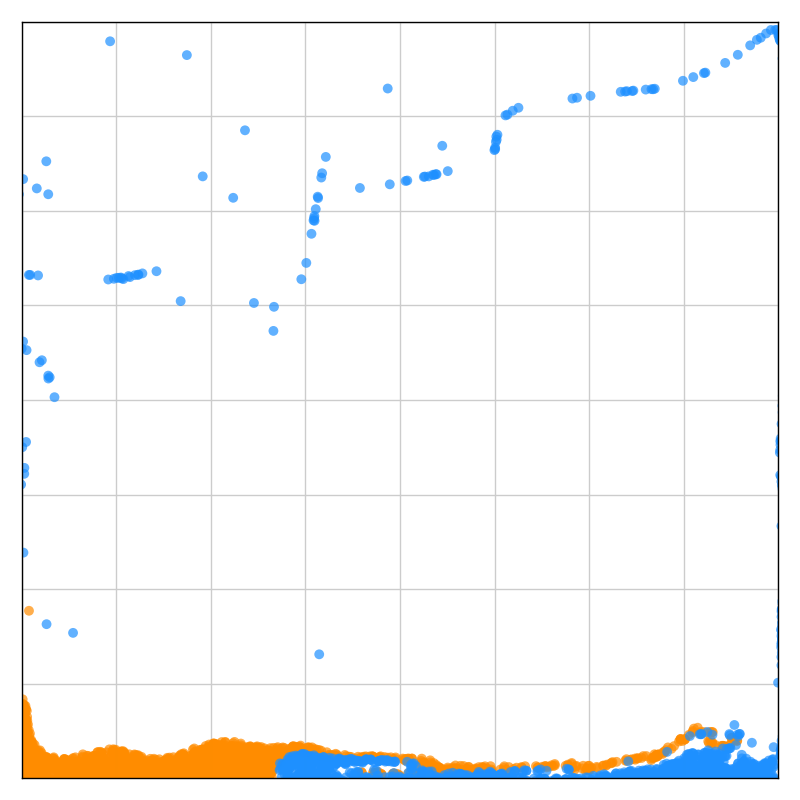}
	\includegraphics[width=0.13\textwidth]{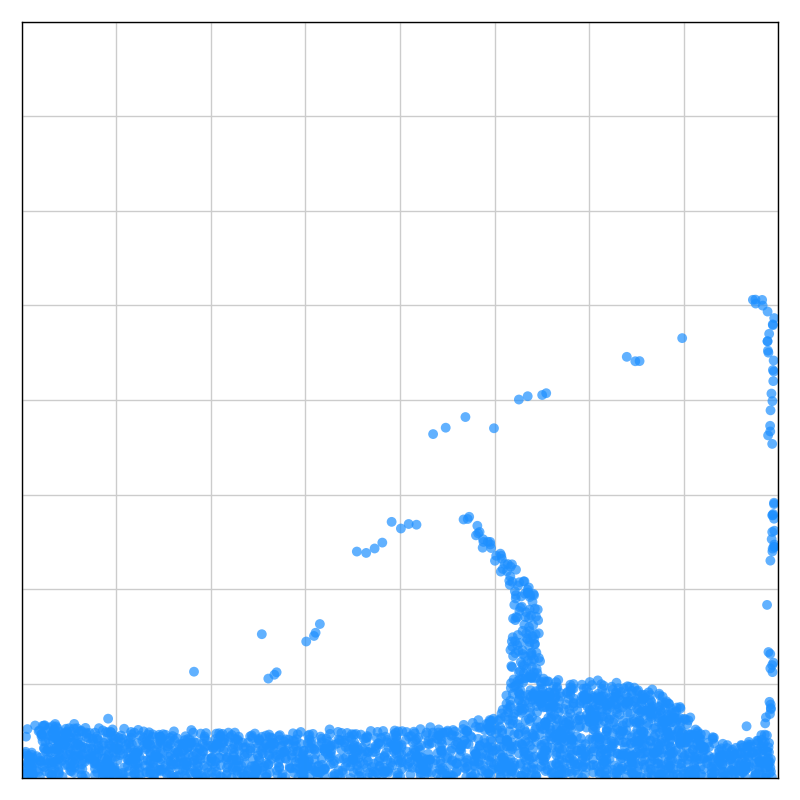}
	\includegraphics[width=0.13\textwidth]{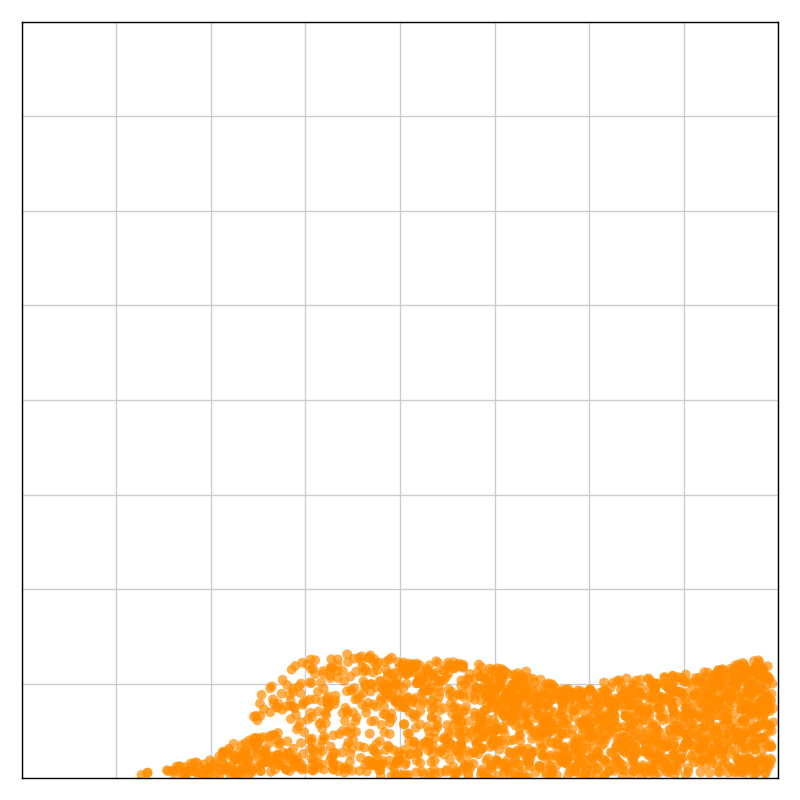}
	\includegraphics[width=0.13\textwidth]{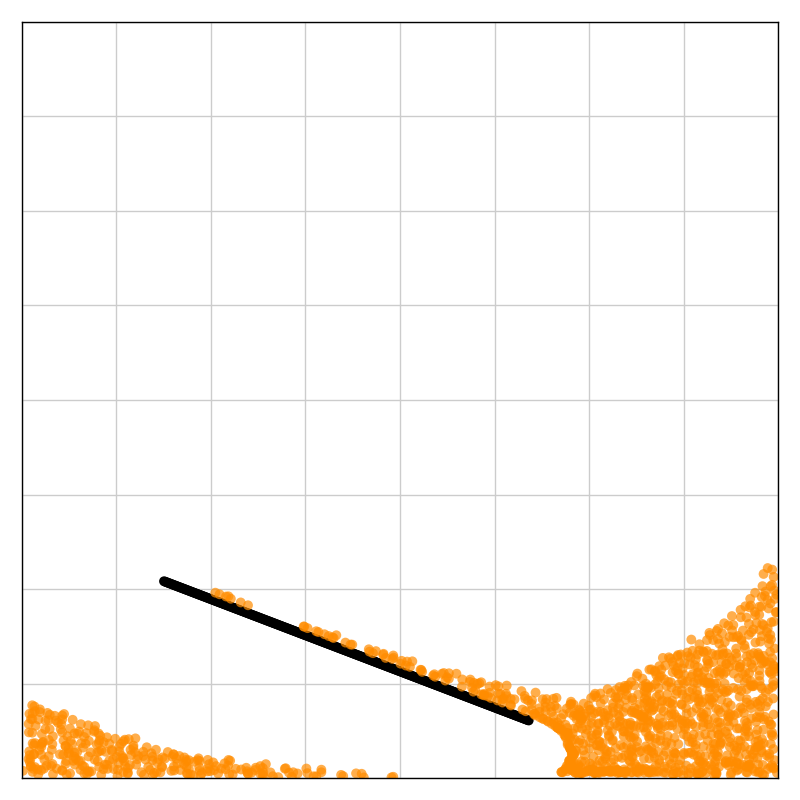}
	\includegraphics[width=0.13\textwidth]{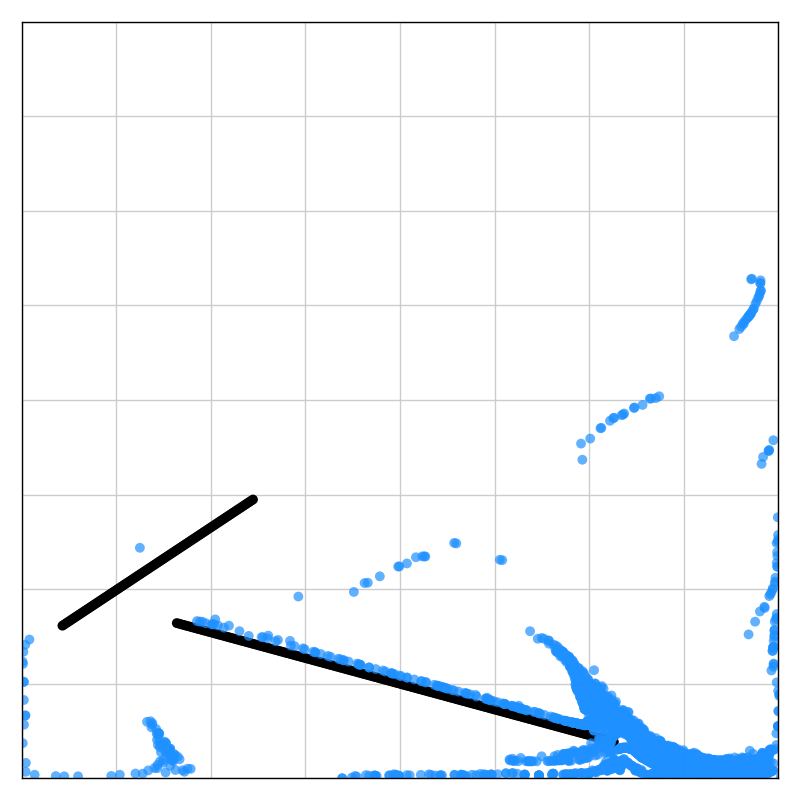}
	\includegraphics[width=0.13\textwidth]{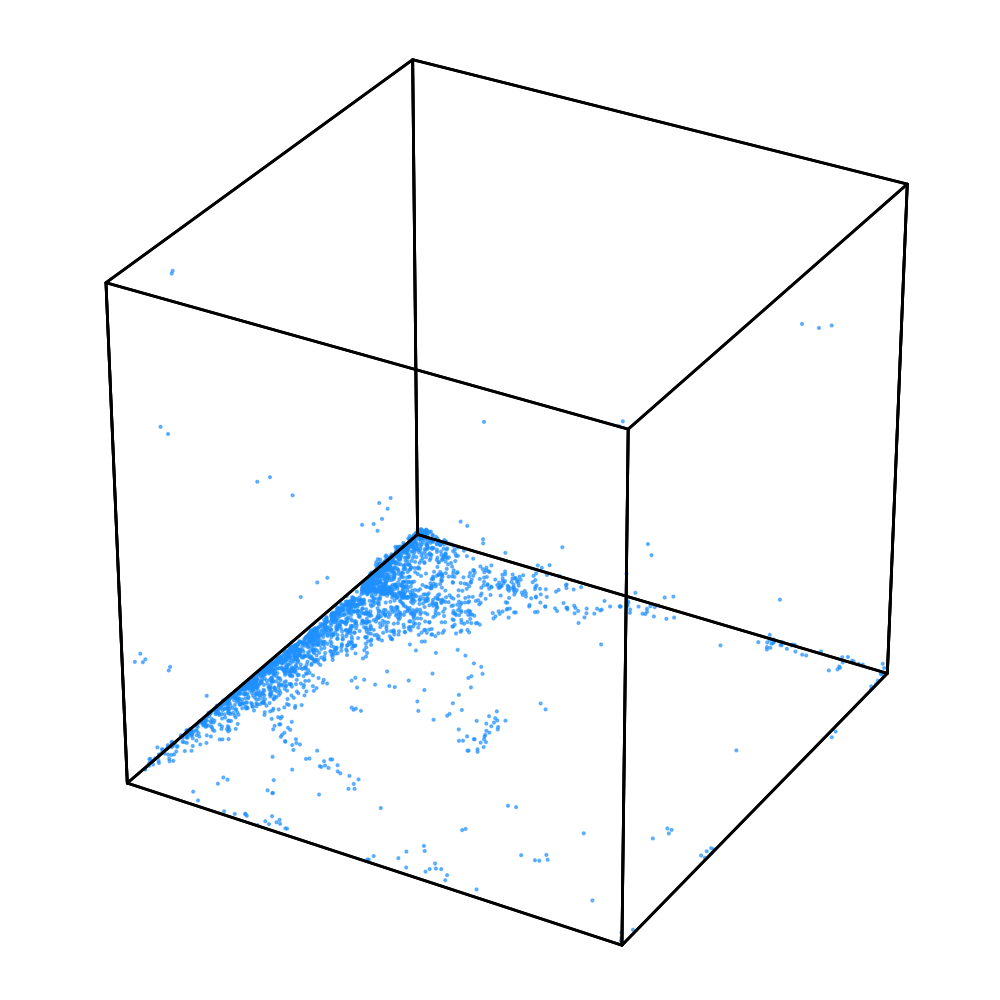}
	\includegraphics[width=0.13\textwidth]{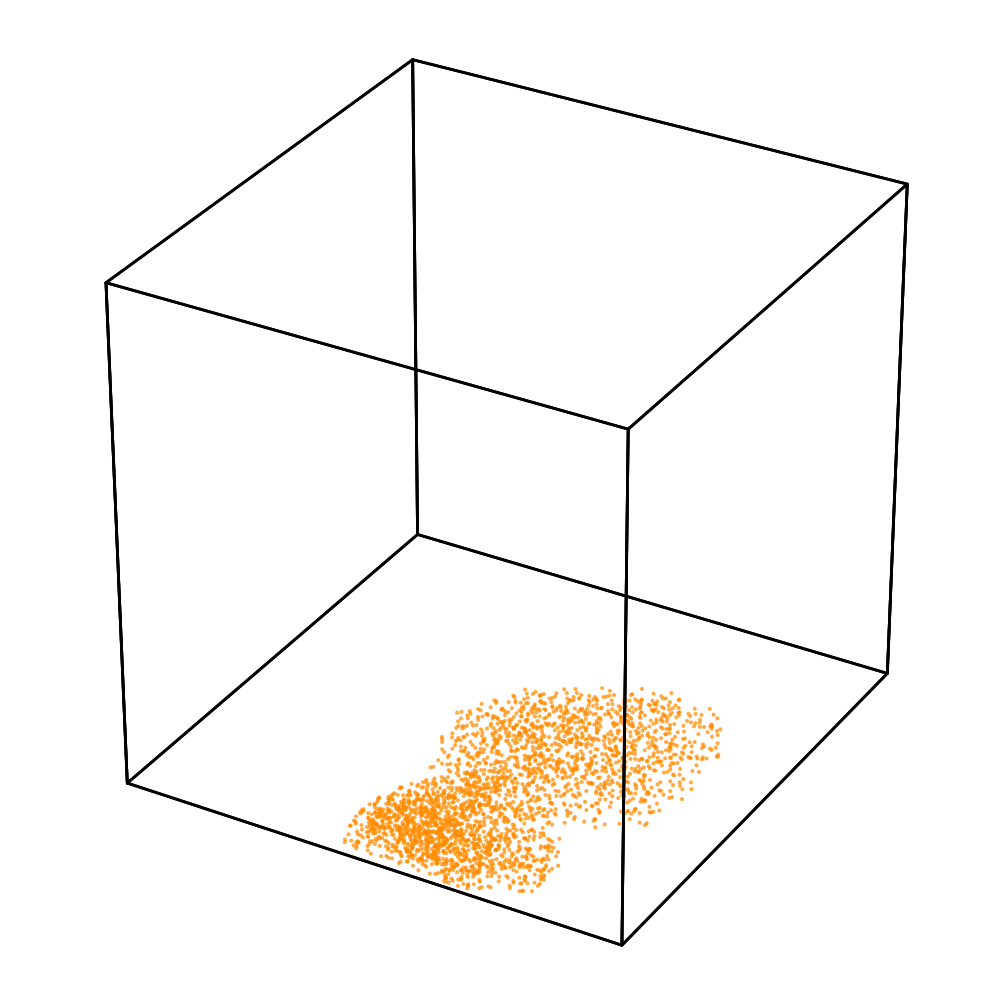}
	\includegraphics[width=0.13\textwidth]{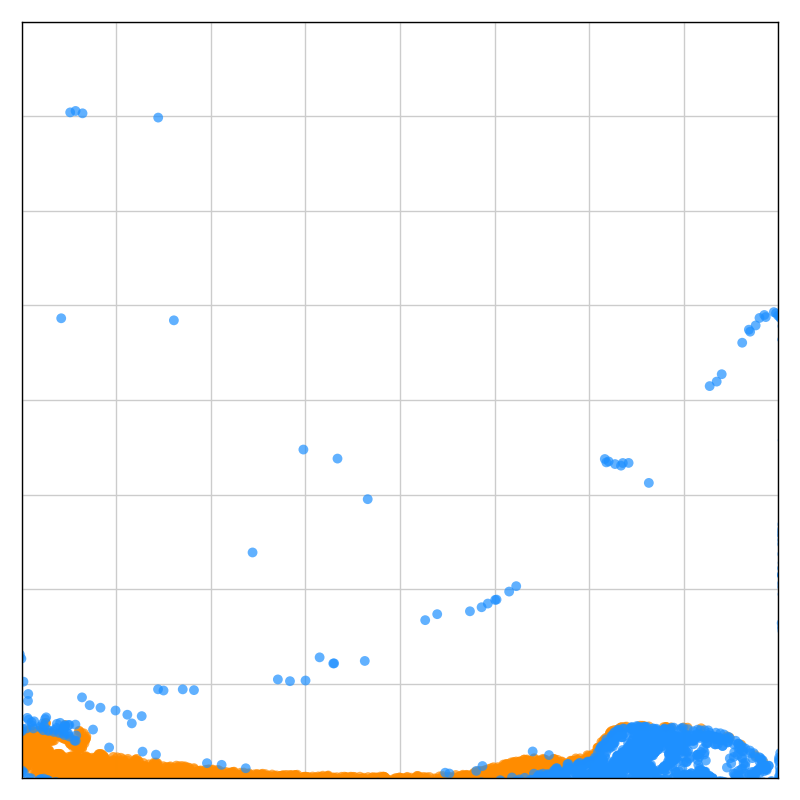}
	\includegraphics[width=0.13\textwidth]{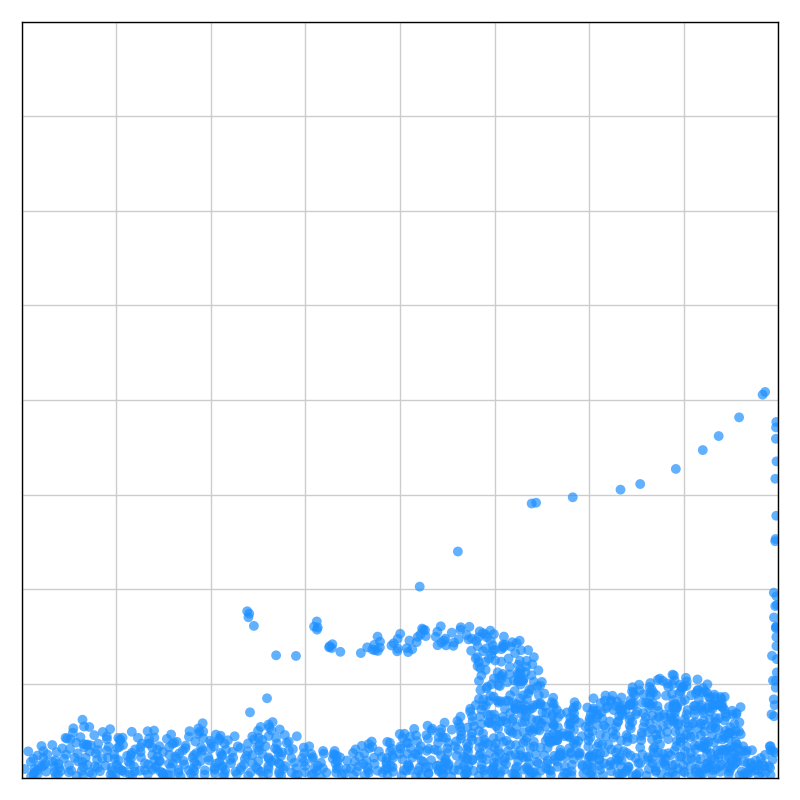}
	\includegraphics[width=0.13\textwidth]{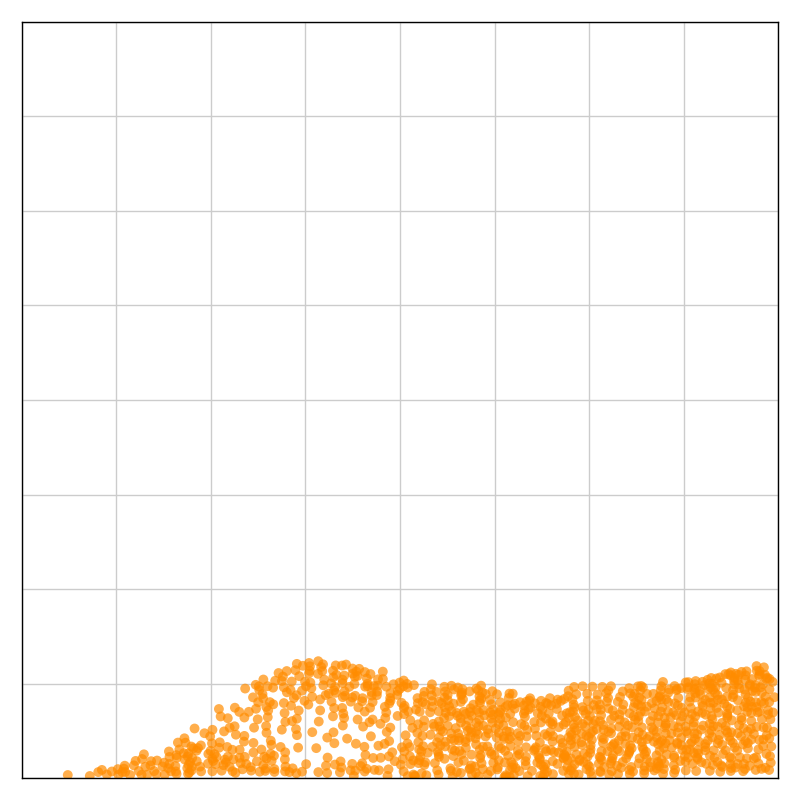}
	\includegraphics[width=0.13\textwidth]{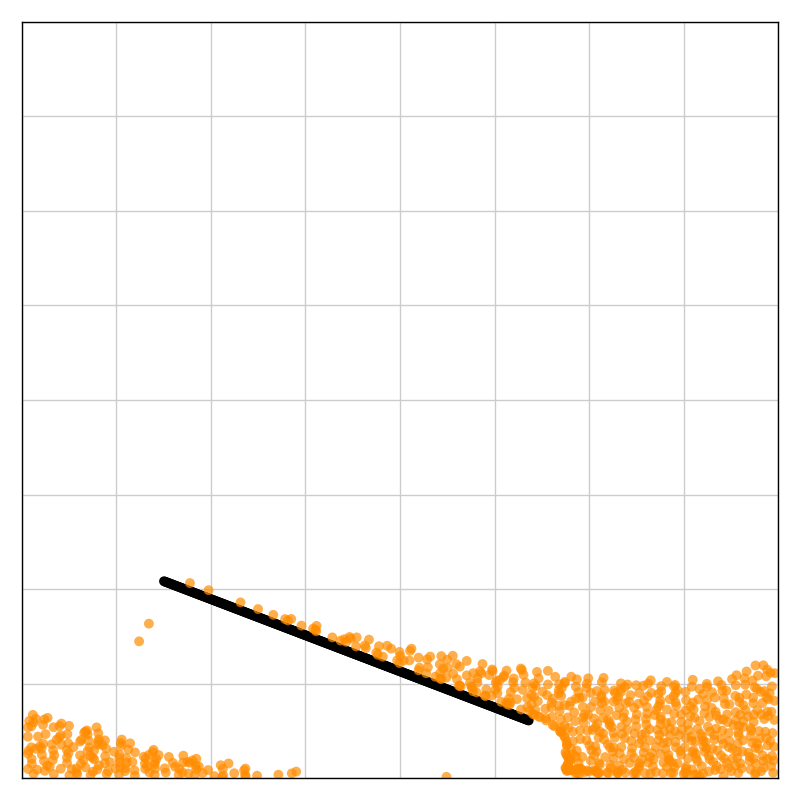}
	\includegraphics[width=0.13\textwidth]{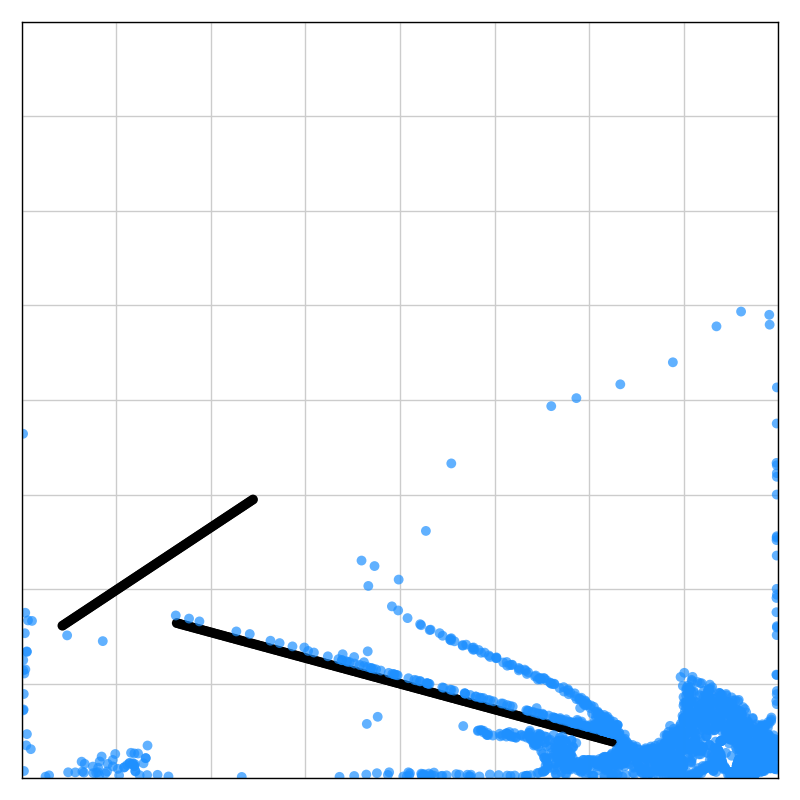}
	\includegraphics[width=0.13\textwidth]{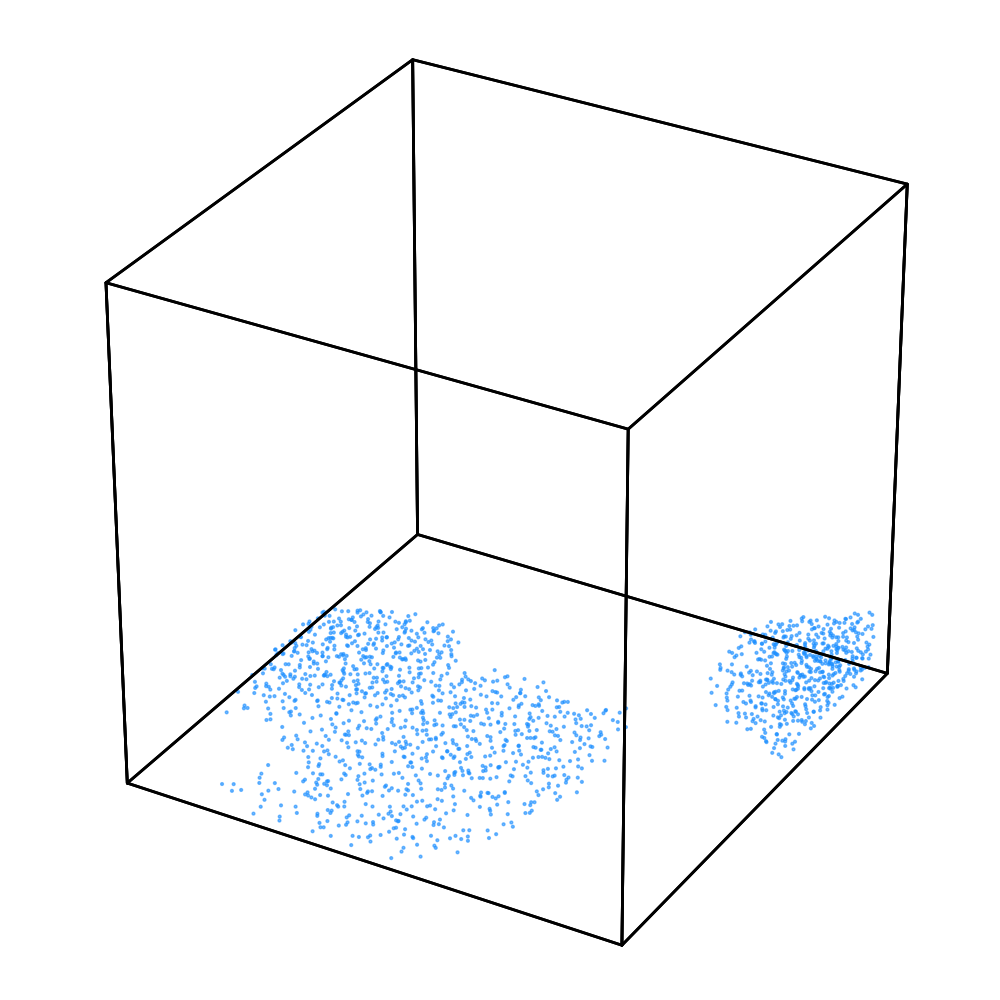}
	\includegraphics[width=0.13\textwidth]{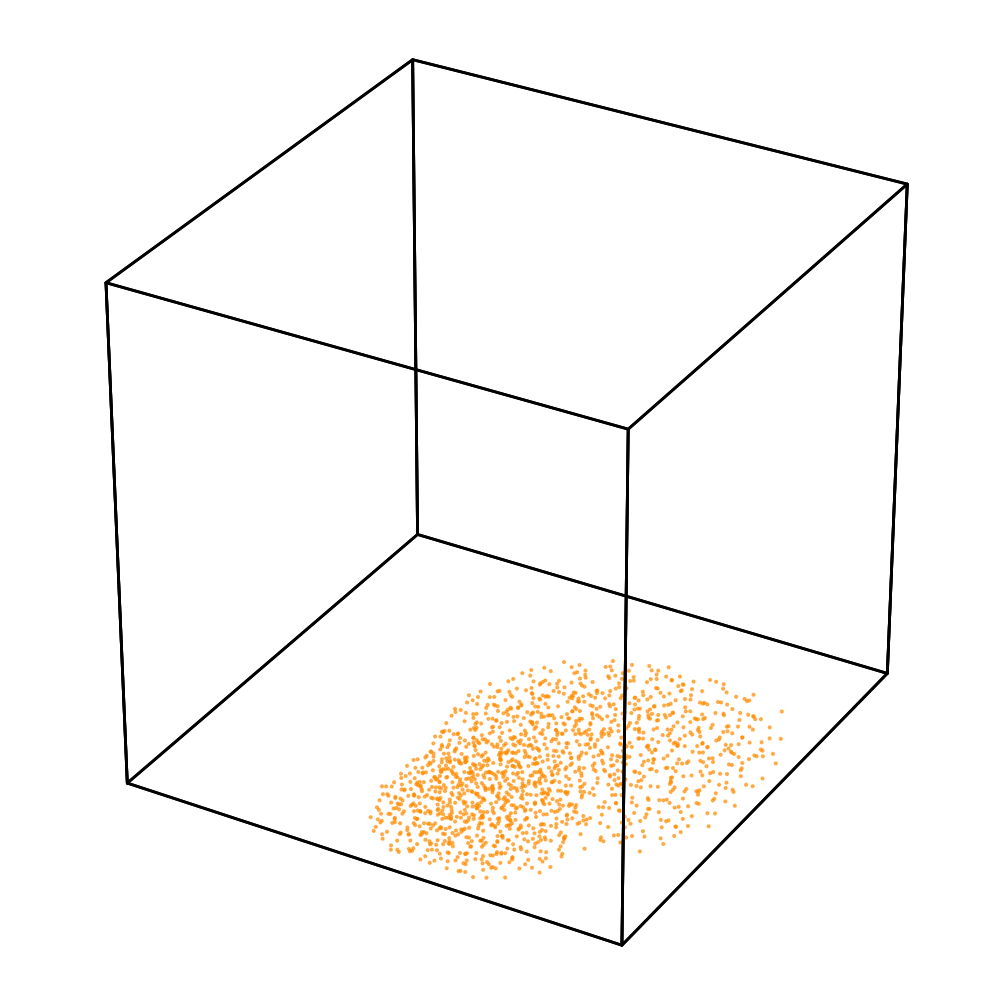}
	\includegraphics[width=0.13\textwidth]{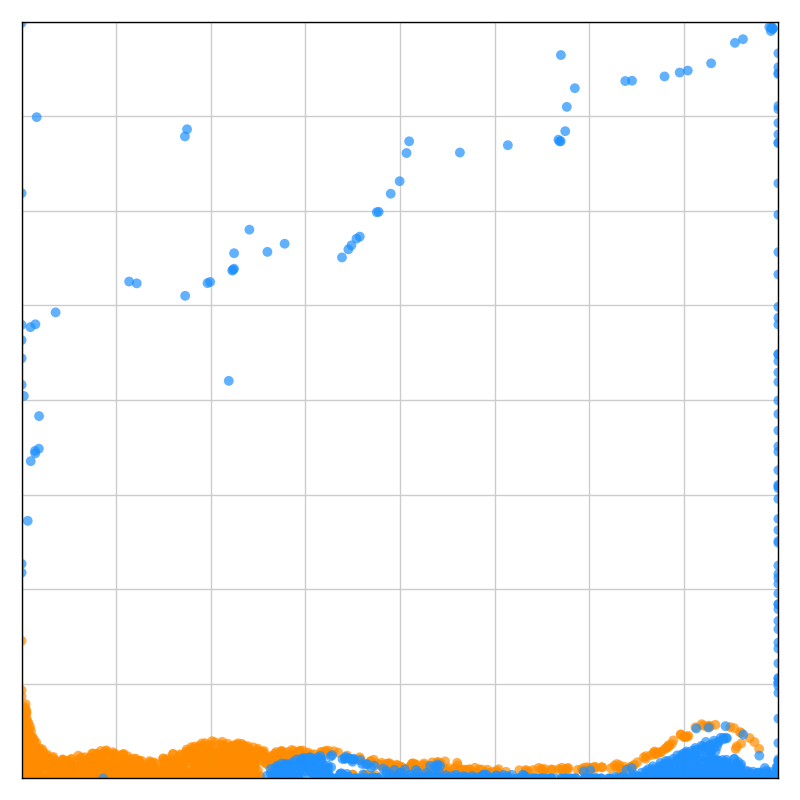}
	\includegraphics[width=0.13\textwidth]{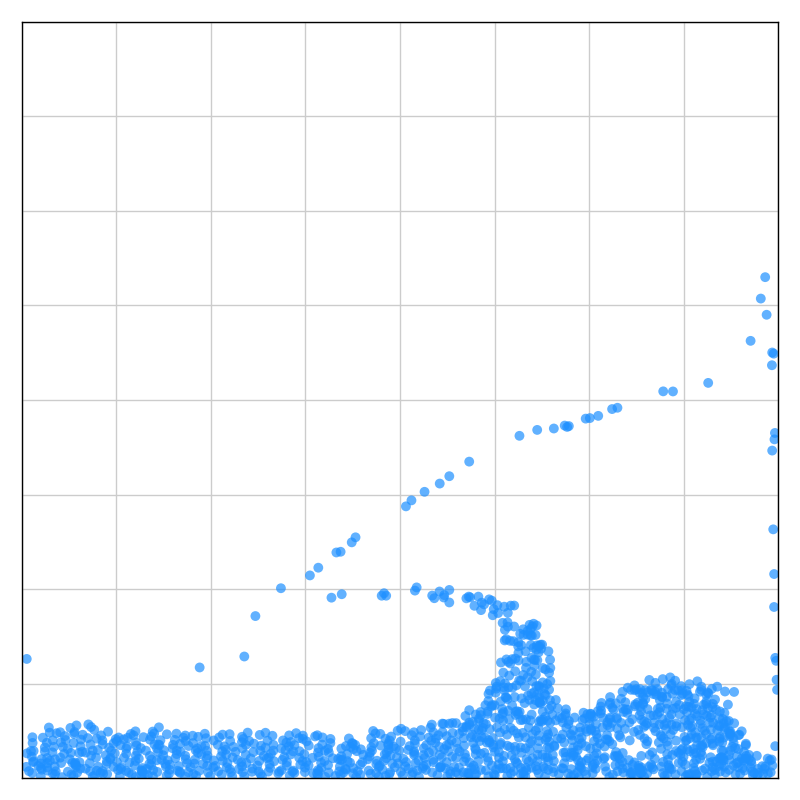}
	\includegraphics[width=0.13\textwidth]{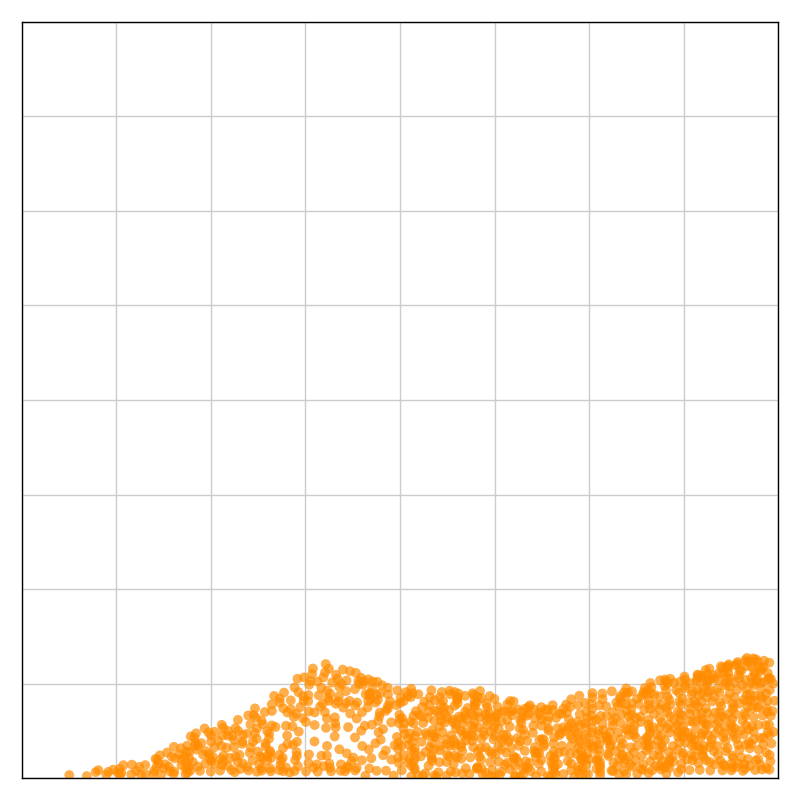}
	\includegraphics[width=0.13\textwidth]{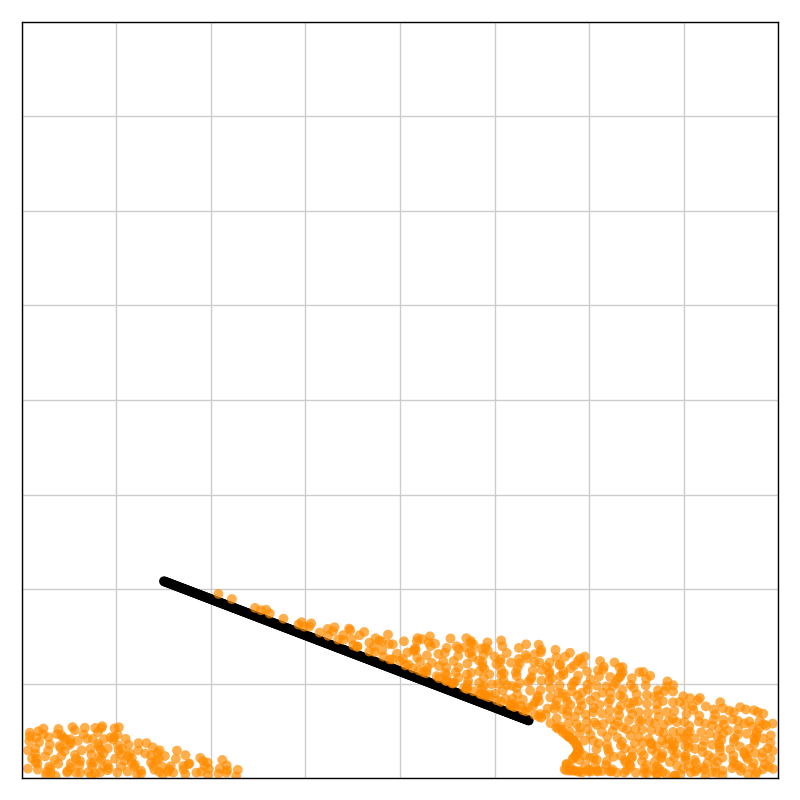}
	\includegraphics[width=0.13\textwidth]{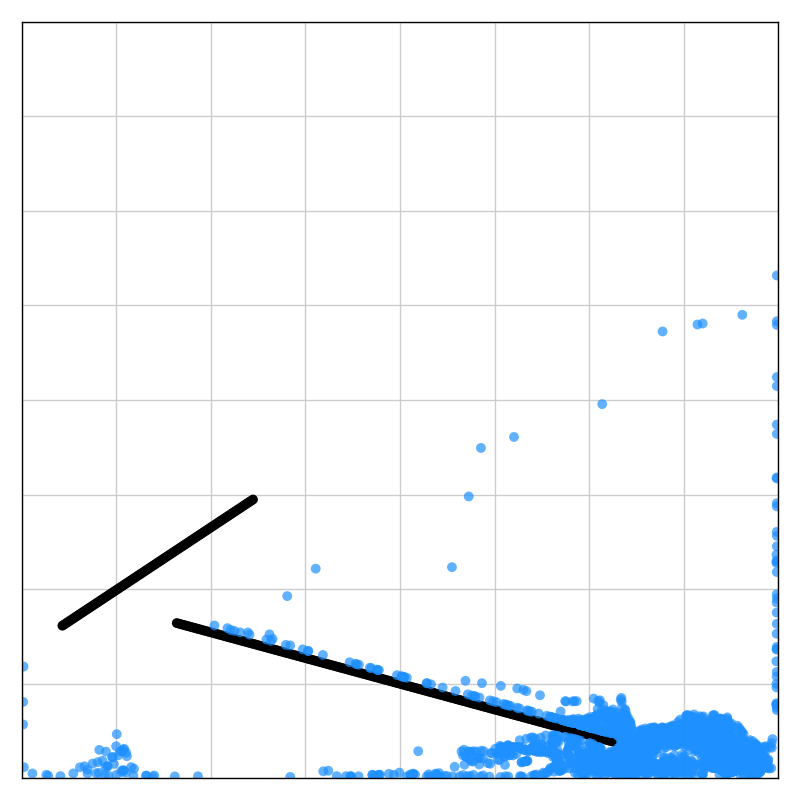}
	\includegraphics[width=0.13\textwidth]{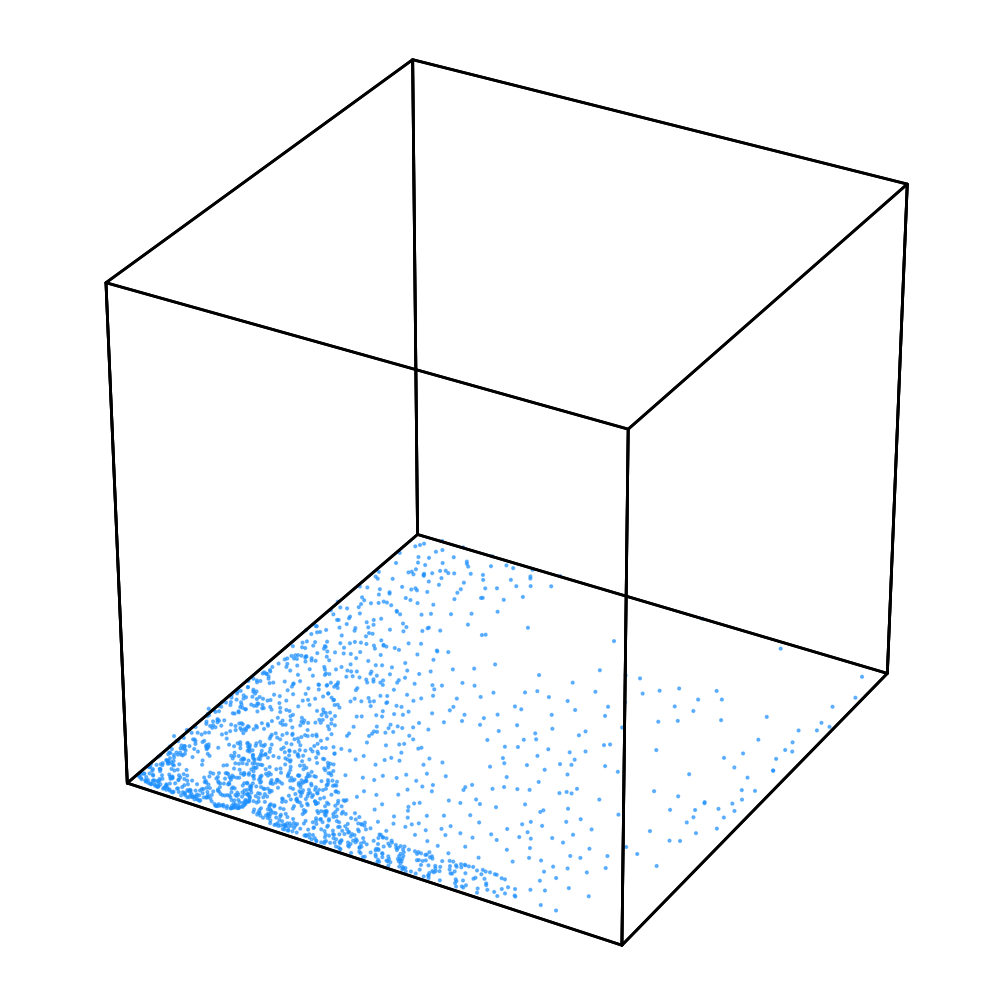}
	\includegraphics[width=0.13\textwidth]{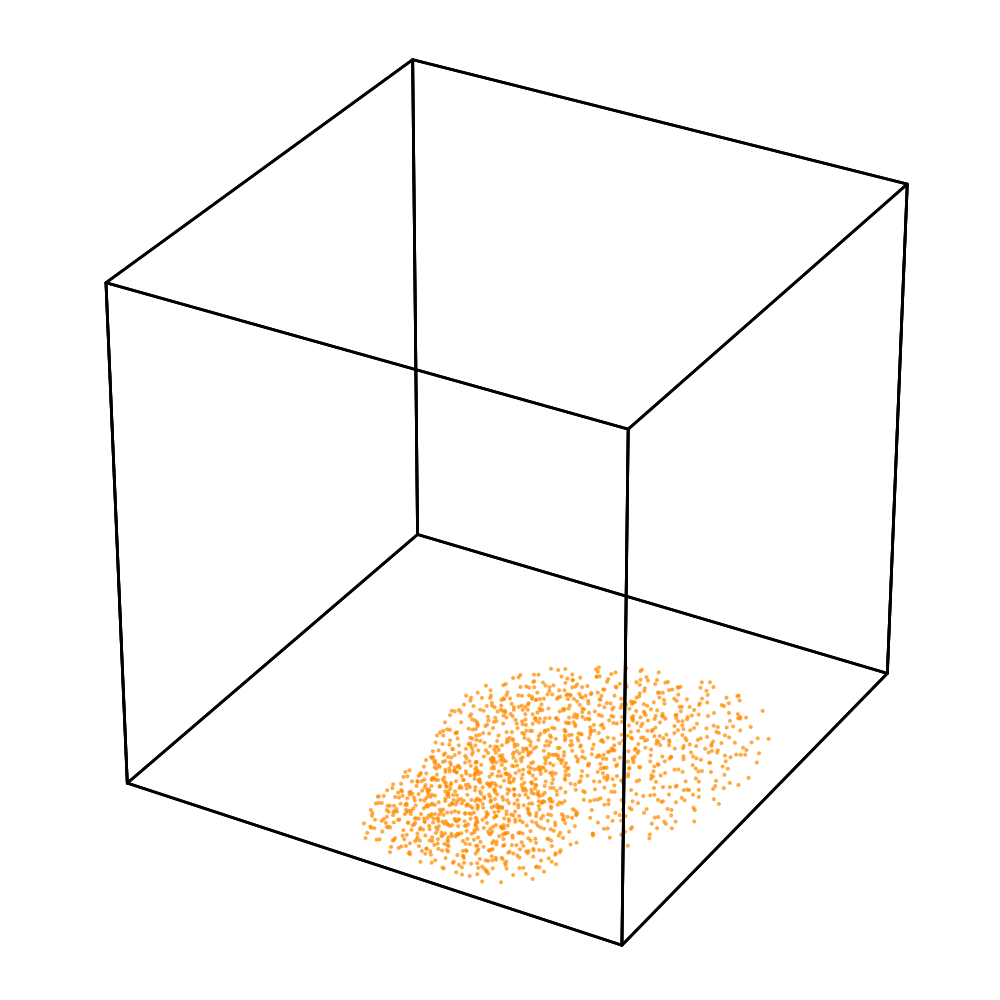}
	\includegraphics[width=0.13\textwidth]{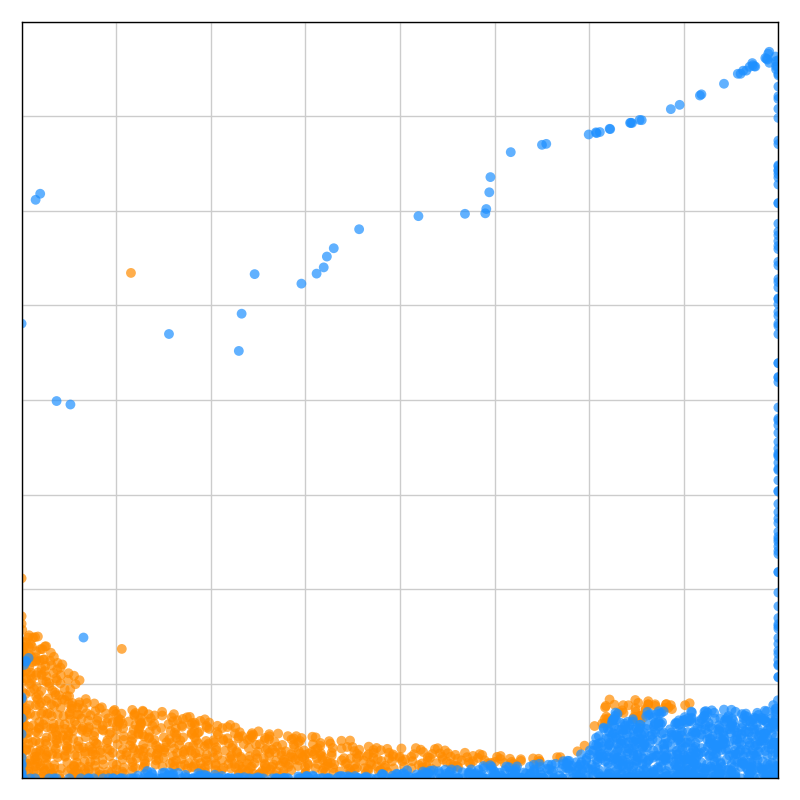}
    \vspace{-0.5em}
    \captionsetup{font=small}
    \caption{Visualizations of fluid simulations by different methods, over different scenarios. From left to right: Water (2D), Sand (2D), SandRamps (2D), WaterRamps (2D), Water (3D), Sand (3D), Water-Sand (2D). From top to bottom: Initial, MPM (ground truth), Original Neural Physics, MPM ($r_p = 1/1.75$), Our Hybrid Solver.}
    \label{fig:fluid_visualizations_more}
    \vspace{-1em}
\end{figure}

\paragraph{Fluid Simulations.}
Figure~\ref{fig:fluid_visualizations_more} presents the visualizations of all models discussed throughout the paper. Here, we show a comparison of intermediate frames from a single trajectory. It is evident that, due to the hybrid design of our hybrid solver, our method produces visual results that are more similar to MPM ($r_p = 1/1.75$) simulations. Since MPM ($r_p = 1/1.75$) is highly consistent with MPM (ground truth), the outputs of our Hybrid solver also align better with MPM compared to the original neural physics.
This demonstrates that our approach effectively balances computational efficiency and accuracy.

\begin{figure}
\centering
	\includegraphics[width=0.8\textwidth]{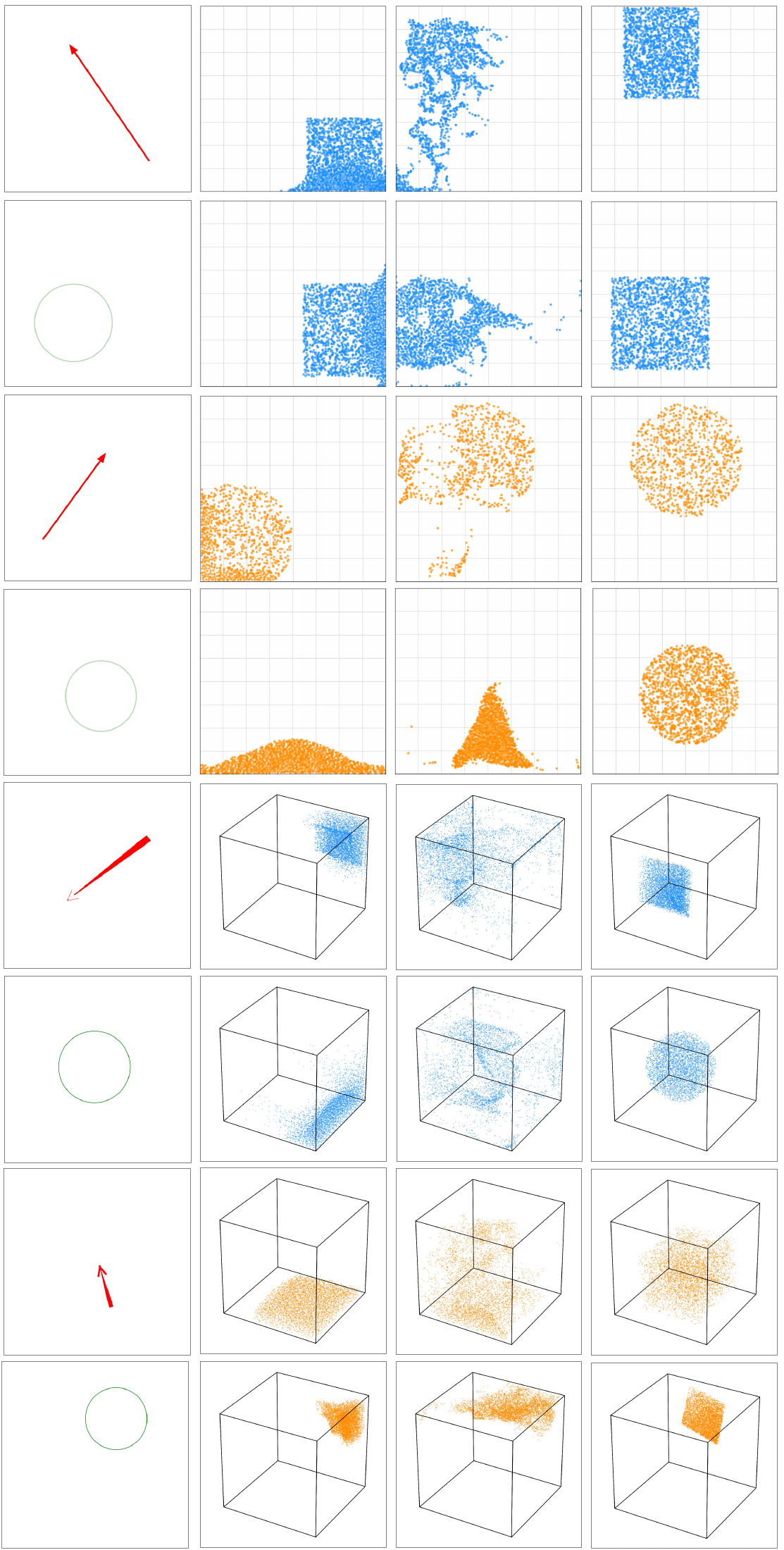}
    \vspace{-0.5em}
    \captionsetup{font=small}
	\caption{More visualization of generative fluid control.
    From top to bottom: Water2D, Sand2D, Water3D, and Sand3D, each with two types of control signals (arrows for motion direction, and oval shapes for target spatial positions).
    From left to right: control signal, initial, ours, ground truth.}
    \label{fig:control_demos_more}
    \vspace{-.5em}
\end{figure}

\paragraph{More Visualizations of Fluid Control.}
Figure~\ref{fig:control_demos_more} presents additional visualizations of generative fluid control across a variety of tasks, both 2D and 3D control signals.
We can see that our approach consistently generates physically plausible and visually accurate outcomes that align closely with the target controls across all fluid types and dimensions, demonstrating strong control capability. These results further confirm the effectiveness of our method in achieving both visually appearing and physically plausible fluid control. 

\end{document}